\documentclass{article}
\usepackage[margin=2.5cm]{geometry}
\usepackage{microtype}
\usepackage{graphicx}
\usepackage{subfig}
\usepackage{booktabs} 
\usepackage{hyperref}
\usepackage{amsmath, amssymb,graphicx,bbm,bbm,color,algorithm2e,dsfont}

\newcommand{\cA}{\mathcal{A}}
\newcommand{\cB}{\mathcal{B}}
\newcommand{\cC}{\mathcal{C}}

\newcommand{\cI}{\mathcal{I}}

\newcommand{\cL}{\mathcal{L}}

\newcommand{\cO}{\mathcal{O}}

\newcommand{\cV}{\mathcal{V}}

\newcommand{\cX}{\mathcal{X}}

\newcommand{\PP}{\mathbb{P}}
\newcommand{\NN}{\mathbb{N}}
\newcommand{\EE}{\mathbb{E}}
\newcommand{\RR}{\mathbb{R}}
\newcommand{\indic}{{\bf 1}}

\newtheorem{proposition}{Proposition}
\newtheorem{theorem}{Theorem}[section]

\newtheorem{definition}[theorem]{Definition}
\newtheorem{claim}[theorem]{Claim}
\begin{document}
\author{Thibaut Cuvelier, Richard Combes, Eric Gourdin}
\title{Statistically Efficient, Polynomial Time Algorithms for Combinatorial Semi Bandits}
\maketitle
\begin{abstract}
We consider combinatorial semi-bandits over a set of arms ${\cal X} \subset \{0,1\}^d$ where rewards are uncorrelated across items. For this problem, the algorithm ESCB yields the smallest known regret bound $R(T) = {\cal O}\Big( {d (\ln m)^2 (\ln T) \over \Delta_{min} }\Big)$, but it has computational complexity ${\cal O}(|{\cal X}|)$ which is typically exponential in $d$, and cannot be used in large dimensions. We propose the first algorithm which is both computationally and statistically efficient for this problem with regret $R(T) = {\cal O} \Big({d (\ln m)^2 (\ln T)\over \Delta_{min} }\Big)$ and computational complexity ${\cal O}(T {\bf poly}(d))$. Our approach involves carefully designing an approximate version of ESCB with the same regret guarantees, showing that this approximate algorithm can be implemented in time ${\cal O}(T {\bf poly}(d))$ by repeatedly maximizing a linear function over $\cX$ subject to a linear budget constraint, and showing how to solve this maximization problems efficiently.
\end{abstract}

\section{Introduction}
	We consider the combinatorial bandit problem with semi-bandit feedback and independent rewards across items. Time is discrete, and at times $t=1,...,T$ a learner chooses a decision $x(t) \in \cX$, where $\cX \subset \{0,1\}^d$ is a combinatorial set that is known to the learner. The learner then receives a reward $Z^{\top}(t) x(t)$ and observes a feedback vector $Y(t) = (x_1(t) Z_1(t),\dots,x_d(t) Z_d(t))$, where $Z(t) \in [0,1]^d$ is a random vector with mean $\theta \in \RR^d$ and whose entries are independent.

	The expected reward from decision $x \in \cX$ is $\theta^\top x$, and the goal is to maximize the sum of expected rewards, or equivalently to minimize the regret:
	\begin{align*}
		R(T) = \sum_{t=1}^T \max_{x \in \cX} \{ \theta^\top x \} - \EE( \theta^\top x(t)).
	\end{align*}
	The vector $\theta$ is unknown to the learner; in order to minimize the regret, one must discover the decision $x \in \cX$ maximizing $\theta^\top x$, and in turn one must explore enough decisions to obtain sufficient information about $\theta$. This problem models a large amount of practically relevant online decision problems such as online shortest path routing, ad-display optimization, and resource allocation.

    \paragraph{Contribution.}

	In this paper, we propose the first (to the best of our knowledge) algorithm with regret $R(T) = {\cal O} \Big({d (\ln m)^2 (\ln T)\over \Delta_{\min}}\Big)$ and polynomial computational complexity in the problem dimension $d$ for a large family of combinatorial sets $\cX$.

	The rest of this paper is organized as follows. In Section~\ref{sec:combinatorial_bandits}, we further highlight the model and describe combinatorial sets of interest. In Section~\ref{sec:related_work}, we outline the related work on this problem, including state-of-the-art regret bounds and algorithms, and highlight our contribution. In Section~\ref{sec:escb}, we describe the proposed algorithm, and provide regret bounds. In Section~\ref{sec:polynomial_time}, we show that our algorithm may be implemented in polynomial time for a large class of combinatorial sets, and analyze its computational complexity in details. In Section~\ref{sec:numerical}, we perform numerical experiments to complement. Section~\ref{sec:conclusion} concludes the paper.

\section{Combinatorial Semi-Bandits}\label{sec:combinatorial_bandits}

    \subsection{Model}
    	As said previously, we consider the following problem. Time is discrete, and at times $t=1,...,T$ a learner chooses a decision $x(t) \in \cX$, where $\cX \subset \{0,1\}^d$ is a combinatorial set which is known to the learner. Set $\cX$ may be any combinatorial set, including the bases of a matroid, the set of paths in some graph, the set of matchings in a bipartite graph, etc. The problem dimension is $d$, and we define $m = \max_{x \in \cX} \indic^{\top} x$ the size of the largest decision.

    	After selecting decision $x(t)$, the learner then receives a reward $Z^{\top}(t) x(t)$ and observes a feedback vector $Y(t) = (x_1(t) Z_1(t),\dots,x_d(t) Z_d(t))$, where $Z(t) \in [0,1]^d$ is a random vector.

    	We assume that $(Z(t))_t$ are i.i.d. with mean $\theta \in [0,1]^d$ and that the entries of $Z_i(t)$ are independent as well. Vector $\theta$ is initially unknown to the learner, and must be learnt by repetitively selecting decisions and observing subsequent feedback. For $i \in \{1,...,d\}$, if $x_i(t) = 1$, then the learner obtains a noisy realization of $\theta_i$ and nothing otherwise, so that decisions must be carefully selected to obtain a good estimate of $\theta$. This is the "semi-bandit feedback" model.

    	Since $\theta$ is unknown to the learner, decision $x(t)$ must be selected solely as a function of the feedback information available at time $t$, i.e. $Y(t-1),...,Y(1)$.

    	The expected reward received by selecting decision $x \in \cX$ is $\theta^\top x$ (i.e. rewards are linear in the decision), so that $\theta_i$ represents the amount of reward received by selecting $x_i = 1$. The optimal decision is $x^\star \in \arg\max_{x \in \cX} \{\theta^\top x\}$ (there may be several optimal decisions). We define the reward gap $\Delta_{x} = \theta^\top(x^\star - x)$, i.e. the amount of regret incurred to the learner by selecting decision $x$ instead of $x^\star$. We denote by $\Delta_{\min} = \min_{x: \Delta_{x} > 0} \Delta_x$ the smallest non-null gap.

    	The goal of the learner is to minimize the regret, which is simply the difference in terms of expected cumulative rewards between the learner and an oracle who knows the latent vector $\theta$ in advance and who always selects the optimal decision $x^\star$, that is:
    	\begin{align*}
    		R(T) = \sum_{t=1}^T \EE( \Delta_{x(t)}).
    	\end{align*}

    \subsection{Combinatorial Sets of Interest}

    	Of course, in order to devise algorithms with low regret and computational complexity, one must take into account the structure of the combinatorial set $\cX$. We will consider several family of sets, which we present here. We denote by $e_i$ for $i=1,...,d$ the $i$-th canonical basis vector. When working with graphs, we identify sets of edges with binary vectors, namely given a graph $G = (V,E)$, we identify a subset of edges $E' \subset E$ with a binary vector $x = \{0,1\}^{|E|}$ where $x_e = 1$ if $e \in E'$ and $x_e = 0$ otherwise.

        These sets all correspond to "easy" combinatorial problems: for most of them, it is possible to optimize exactly a linear function in polynomial time. The exception is the knapsack, which can be approximated in polynomial time. They also have a practical significance. Source-destination paths correspond to the online shortest-path-routing problem; $m$-sets model ad-display optimization; matchings can be used to perform resource allocation in wireless networks.

        \paragraph{$m$-sets.}
            The set of $m$ sets is the set of vectors $x = \{0,1\}^d$ which have at most $m$ non-null entries, namely:
            \begin{align*}
            	\cX = \{x \in \{0,1\}^d : (1,...,1)^\top x  \le m\}.
            \end{align*}

        \paragraph{Spanning trees.}
            Consider a graph $G = (V,E)$. A spanning tree $x \in \{0,1\}^{|E|}$ is a subset of edges which covers each vertex $v \in V$ at least once and forms a tree. The dimension here is the number of edges: $d=|E|$.
            \begin{align*}
            	\cX = \{x \in \{0,1\}^{|E|} : \sum_{e: v \in e} x_e \ge 1 \forall v \text{ and } x \text{ is a tree}\}.
            \end{align*}

        \paragraph{Matroids.}
            A matroid $\cI$ over a set with $d$ elements is a set of vectors $x \in \{0,1\}^d$ which verify two properties: $(i)$ the inclusion property: if $x \le x'$ and $x' \in \cI$ then $x \in I$ and $(ii)$ the exchange property: if $x \in \cI$, $x' \in \cI$ and $\indic^\top x < \indic^\top x'$, then there exists an index $i$ such that $x_i = 0$, $x'_i = 1$, and $x + e_i \in \cI$. Since $m$-sets and spanning trees of a graph form a matroid, any algorithm for matroids also applies to $m$-sets and spanning trees.

        \paragraph{Source-destination paths.}
            Consider $G = (V,E)$ a directed acyclic graph and $u$ and $v$ in $V$ two vertices. The set of paths between source $u$ and destination $v$ is defined as:
            \begin{align*}
            	\cX = \{x \in \{0,1\}^{|E|}:  \sum_{e \in in(w)} x_e - \sum_{e \in out(w)} x_e
            	            	=  \indic\{w=u\} - \indic\{w=v\}, w \in V\}.
            \end{align*}
            where $in(v)$ and $out(v)$ are the set of incoming and outgoing edges respectively of $v \in V$. The dimension $d = |E|$ is the number of edges and $m = \max_{x \in \cX} \indic^\top x$ is the length of the longest path from $u$ to $v$.

        \paragraph{Matchings.}
            Consider $G = (V,E)$ a bipartite graph. A matching is a set of edges which cover each vertex $v \in V$ at most once, and the set of matchings is:
            \begin{align*}
            	\cX = \{x \in \{0,1\}^{|E|} : \sum_{e \in E: v \in e} x_e \le 1 \forall v \}.
            \end{align*}

        \paragraph{Intersection of two matroids.}
            Consider $\cI$ and $\cI'$ two matroids over a common set with $d$ elements, their intersection is:
            \begin{align*}
            	\cX = \{x \in \{0,1\}^{d} : x \in \cI, x \in \cI'\}.
            \end{align*}
            Since the set of matchings of any bipartite graph is the intersection of two matroids, algorithms for the intersection of two matroids also apply to matchings of a bipartite graph.

        \paragraph{Knapsack-like sets.}
            Consider $A \in \NN^{d \times k}$ matrix and $c \in \NN^{k}$ a vector both with positive integer entries. The corresponding knapsack set is:
            \begin{align*}
            	\cX = \{x \in \{0,1\}^{d} : A x \le c\}.
            \end{align*}
            We call $\cX$ a knapsack-like set as its elements $x$ verify $k$ knapsack constraints $\sum_{i=1}^d A_{\ell,i} x_i \le c_i$, $\ell=1,...,k$. For $k = 1$, $\cX$ is the set of feasible solutions of a knapsack problem and the set of $m$-sets is a knapsack-like set.

    \subsection{Maximization Problems}

    	In order to select a decision $x(t)$ at time $t$, most known algorithms involve maximizing some function over the set of decisions $\cX$. Hence the computational complexity of these algorithms depends mostly on the complexity of these optimization problems. We consider $a,b$ two vectors of $\RR^d$ with positive entries.

    	\paragraph{Linear Maximization.}
            Problem $(P_1)$ involves maximizing a linear function over $\cX$: 
            \begin{align*}
            	\underset{x}{\max}  \{a^\top x \} \text{  subject to } x \in \cX \quad (P_1)
            \end{align*}

    	\paragraph{Index Maximization.}
            Problem $(P_2)$ involves maximizing the sum of a linear function and the square root of a linear function over $\cX$: 
            \begin{align*}
            	\underset{x}{\max} \{ a^\top x + \sqrt{b^\top x} \} \text{  subject to }  x \in \cX \quad (P_2)
            \end{align*}

    	\paragraph{Budgeted Linear Maximization.}
            Problem $(P_3)$ involves maximizing a linear function over $\cX$ subject to a linear budget constraint: 
            \begin{align*}
            	\underset{x}{\max} \{ b^\top x \} \text{ subject to } x \in \cX \text{ and } a^\top x \ge s \quad (P_3)
            \end{align*}

\section{Related Work and Contribution}\label{sec:related_work}
	The study of classical bandits dates back to~\cite{robbins1952} and~\cite{lai1985}. The order optimal regret in this problem is $R(T) = {\cal O}\Big({d (\ln T) \over \Delta_{\min} }\Big)$, which is attained by algorithms such as UCB1~\cite{auer2003} and KL-UCB~\cite{cappe2012}. Linear bandits extend classical bandits when the expected reward is a linear function of the decision~\cite{dani08}. When the set of decisions is a combinatorial set, we have a combinatorial bandit which comes in two version: full-bandit feedback and semi-bandit feedback~\cite{cesa2012}.

	We consider combinatorial semi-bandits with independent rewards across items. For such problems, the first lower bound is due to~\cite{kveton2014} and is specific to matroid bandits, a very specific kind of combinatorial bandit. The best known regret bounds are $R(T) = {\cal O}\Big({d m (\ln T) \over \Delta_{\min} }\Big)$ for CUCB~\cite{chen2013,kveton2014tight} and Thompson sampling~\cite{wen2015,wang2018} \footnote{The authors of \cite{wang2018} study a slightly more general problem; in appendix, we show that their regret bound reduces to $R(T) = {\cal O}\Big({d m (\ln T) \over \Delta_{\min} }\Big)$ in the problem we study.}. Both these algorithms enjoy a polynomial time complexity, as they only solve optimization problem $(P_1)$. The ESCB algorithm was proposed in~\cite{combes2015} where the authors prove a $R(T) = {\cal O}\Big({d \sqrt{m} (\ln T) \over \Delta_{\min} }\Big)$ regret bound. As a follow-up, \cite{degenne2016} propose OLS-UCB which, in our problem, reduces to ESCB, and thereby prove that ESCB in fact achieves $R(T) = {\cal O}\Big({d (\ln m)^2 (\ln T) \over \Delta_{\min} }\Big)$. Therefore, ESCB achieves the best known regret bound for the problem at hand; however, its computational complexity is not polynomial in the dimension. Namely, ESCB must, at each step, solve optimization problem $(P_2)$, which is in general $\mathcal{NP}$-hard~\cite{atamturk2017}. For $m$-sets, paths, and matchings, $P_2$ is $\mathcal{NP}$-hard, see Section~\ref{P2nohard} of the appendix for a more in-depth discussion. There seems to be an interesting interplay between statistical efficiency (regret) and computational complexity, which is summarized in Table~\ref{regret_complexity}.

	We also highlight that there exist algorithms for particular combinatorial semi-bandits (for $m$-sets, spanning trees, and more generally matroids) with both polynomial complexity and order optimal regret $R(T) = {\cal O}\Big({d (\ln T) \over \Delta_{\min} }\Big)$, see \cite{anantharam1987asymptotically_iid,talebi2016,perrault2019}. However, those algorithms do not extend to general combinatorial bandits, unlike ESCB, and to some extent AESCB.

	Much less is known about combinatorial bandits with full-bandit feedback, which is a harder problem. Of course, combinatorial bandits with full-bandit feedback are a particular case of linear bandits, so that algorithms from \cite{dani08,abbasi11,chu11} can be applied. The state of the art is \cite{rejwan20}, which proposes an algorithm with regret $R(T) = {\cal O}\Big({d m(\ln T) \over \Delta_{\min} }\Big)$. However, this algorithm only works for $m$-sets, and whether or not the same regret can be achieved for any combinatorial set $\cX$ is an open problem.

    \paragraph{Our Main Contribution.}
        We propose AESCB (approximate ESCB), the first algorithm that is both computationally and statistically efficient for this problem: it has a state-of-the-art regret bound and a polynomial time complexity. We show that, whenever budgeted linear maximization over $\cX$ can be solved up to a given approximation ratio, AESCB is implementable in polynomial time. More precisely, its regret is $R(T) = {\cal O} \Big({d (\ln m)^2 (\ln T)\over \Delta_{\min} }\Big)$ and its complexity is ${\cal O}(  {\bf poly}(d,\delta_{T}^{-1}))$ for a large class of combinatorial semi-bandits, where $\delta_{t}$ is any function which vanishes as $t \to \infty$, for instance $\delta_t = (\ln t)^{-1}$ or $\delta_t = (\ln \ln t)^{-1}$. We release the code (in Julia~\cite{julia}) of AESCB for others to experiment with it.

    \begin{table}[t]
        \caption{Regret and complexity of algorithms.}
        \label{regret_complexity}
        \vskip 0.15in
        \begin{center}
            \begin{small}
                \begin{sc}
                    \begin{tabular}{lcr}
                        \toprule
                        Algorithm & Best Regret Bound & Complexity \\
                        \midrule
                        CUCB & $\cO\Big( {d m (\ln T) \over \Delta_{\min} }\Big)$ & $\cO({\bf poly}(d))$ \\
                        TS & $\cO\Big( {d m (\ln T) \over \Delta_{\min} }\Big)$ & $\cO({\bf poly}(d))$ \\
                        ESCB & $\cO\Big( {d (\ln m)^2 (\ln T) \over \Delta_{\min} }\Big)$ & $\cO(|\cX|)$ \\
                        AESCB & $\cO\Big( {d (\ln m)^2 (\ln T) \over \Delta_{\min} }\Big)$ & $\cO(\delta_{t}^{-1} {\bf poly}(d))$ \\
                        \bottomrule
                    \end{tabular}
                \end{sc}
            \end{small}
        \end{center}
        \vskip -0.1in
    \end{table}

\begin{table}[t]
    \caption{Complexity of algorithms (at time step $t$) as a function of the chosen discretisation $\delta_t$.}
    \label{complexity_table}
    \vskip 0.15in
    \begin{center}
        \begin{small}
            \begin{sc}
                \begin{tabular}{lccr}
                    \toprule
                    Algorithm & CUCB \& TS & AESCB \\
                    \midrule
                    $m$-sets & $\cO\left( d \ln d \right)$ & $\cO\left(d m \delta_t^{-2}\right)$ \\
                    Spanning Trees & $\cO\left( d \ln d \right)$  & $\cO\left(m d \left(\ln d\right)^{3} \delta_t^{-1} \right)$  \\
                    Matroids & $\cO\left( d \ln d \right)$  & $\cO\left(m d \left(\ln d\right)^{3} \delta_t^{-1} \right)$  \\
                    Paths & $\cO\left(d\ln d\right)$ & $\cO\left(d^2 + d m \delta_t^{-1} \right)$ \\
                    Matchings & $\cO\left(m^3\right)$ & $\cO\left(m d^{10} \delta_t^{-1} \right)$ \\
                    \bottomrule
                \end{tabular}
            \end{sc}
        \end{small}
    \end{center}
    \vskip -0.1in
\end{table}

\section{Exact and Approximate ESCB Algorithms}\label{sec:escb}

	We propose the AESCB algorithm for combinatorial semi-bandits, which is an approximate version of the ESCB algorithm. We prove that AESCB and ESCB both enjoy the same regret bound. In the subsequent sections, we will also show that AESCB, unlike ESCB, can be implemented in polynomial time with respect to the dimension for a large class of combinatorial problems.

    We define the following statistics, for $i=1,...,d$:
    \begin{align*}
    	n_i(t) &= \sum_{t'=1}^{t-1} x_i(t') \\
    	\hat\theta(t) &= {\sum_{t'=1}^{t-1} x_i(t') Z_i(t')  \over  \max(1,\sum_{t'=1}^{t-1} x_i(t'))} \\
    	\sigma^2_i(t) &= \begin{cases} {f(t) \over 2 n_i(t)} & \text{ if } n_i(t) \ge 1 \\ +\infty & \text{otherwise.} \end{cases}
    \end{align*}
    where, at time $t$, $n_i(t)$ is the number of samples obtained for $\theta_i$, $\hat\theta_i(t)$ is the estimate of $\theta_i$, and $\sigma^2_i(t)$ is proportional to the variance of estimate $\hat\theta_i(t)$. $f(t)$ is defined as $\ln t + 4m\ln\ln t$. We denote by $n(t) = (n_i(t))_{i=1,\dots,d}$, $\hat\theta(t) = (\hat\theta_i(t))_{i=1,\dots,d}$, and $\sigma^2(t) = (\sigma^2_i(t))_{i=1,\dots,d}$ the corresponding vectors.

    \subsection{The ESCB Algorithm}
    	\begin{definition}[ESCB] The ESCB algorithm is the policy which at any time $t \ge 1$ selects decision:
    	\begin{align*}
    		x(t) \in \mathop{\arg\max}_{x \in \cX} \{ \hat\theta(t)^\top x + \sqrt{\sigma^2(t)^\top x}   \}
    	\end{align*}
    	where ties are broken arbitrarily.
    	\end{definition}

    	The ESCB algorithm is an optimistic algorithm, where the index $\hat\theta(t)^\top x + \sqrt{\sigma^2(t)^\top x}$ serves as an upper confidence bound of the unknown reward of decision $x$, $\theta^\top x$. The nonlinear term is the estimated standard deviation of the reward, as the random variables are considered to be independent. Also, ESCB is a natural extension of the UCB1 algorithm for classical bandits. In order to implement ESCB, one needs to solve the optimization problem $(P_2)$ at each time step.

    	The regret of ESCB was analyzed by~\cite{combes2015} and then improved by~\cite{degenne2016}. The regret bound is presented in Theorem~\ref{th:escb_regret}, as the algorithm of~\cite{degenne2016} reduces to that of~\cite{combes2015} when rewards are uncorrelated across items.

    	\begin{theorem}[Regret of ESCB] \label{th:escb_regret}
    		The regret of ESCB admits the following upper bound for all $T \ge 1$:
    		\begin{align*}
    			R\!\left(T\right) &\leq C_4(m) + \frac{2 d m^{3}}{\Delta_{\min}^{2}}
                                  + \frac{96  d f(T)}{\Delta_{\min}} \left\lceil \frac{\ln m}{1.61}\right\rceil ^{2}
    		\end{align*}
    		with $f(t) = \ln t+ 4 m \ln\ln t$ and $C_4(m)$ a positive number that solely depends on $m$.

    		By corollary:
   			\begin{align*}
    			R\!\left(T\right)=\mathcal{O}\!\left(d\,(\ln m)^2\,\frac{1}{\Delta_{\min}}\,\ln T\right)\quad\text{as}\quad T\to\infty.
    		\end{align*}
    	\end{theorem}

    	The regret upper bound of Theorem~\ref{th:escb_regret} is the best known regret upper bound for combinatorial semi-bandits with independent rewards across items, so that ESCB is, to the best of our knowledge, the state-of-the art algorithm for this problem in terms of regret. However, ESCB involves solving optimization problem $(P_2)$ at each step, and this problem is $\mathcal{NP}$-hard~\cite{atamturk2017}, so one cannot implement it efficiently as is.

    \subsection{AESCB}
    	We now propose AESCB (Approximate-ESCB), an algorithm that approximates ESCB and enjoys the same regret bound, while being implementable with polynomial complexity. The AESCB algorithm requires two sequences $(\varepsilon_t,\delta_t)$, which quantify the level of approximation at each time step.

    	\begin{definition}[AESCB] The AESCB algorithm with approximation factors $(\varepsilon_t,\delta_t)_{t \ge 1}$ is the policy which at any time $t \ge 1$ selects a decision $x(t)$ verifying:
    	\begin{align*}
    		\mathop{\arg\max}_{x \in \cX} \{ \hat\theta(t)^\top x + \sqrt{\sigma^2(t)^\top x}   \}
    		\le \delta_t + \hat\theta(t)^\top x(t) + {1 \over \varepsilon_t} \sqrt{\sigma^2(t)^\top x(t)}
    	\end{align*}
    	where ties are broken arbitrarily.
    	\end{definition}

    	When $(\varepsilon_t,\delta_t) = (1,0)$ for all $t \ge 1$, AESCB reduces to ESCB. The rationale is that ESCB requires to solve optimization problem $(P_2)$ at each time step, and while $(P_2)$ is $\mathcal{NP}$-hard and cannot be solved exactly in polynomial time (unless $\mathcal{P}=\mathcal{NP}$), it can be approximated in polynomial time in many cases of interest, so that AESCB lends itself to polynomial-time implementation. We show how to do this in Section~\ref{sec:polynomial_time}.

    \subsection{Regret Analysis of AESCB}
        Our first main result is Theorem~\ref{th:aescb_regret}, which provides a regret upper bound for AESCB. We show that, if one chooses approximation parameters $(\varepsilon_t,\delta_t)$ with $\varepsilon_t = \varepsilon>0$ some fixed number and $\delta_t$ any sequence such that $\lim_{t \to \infty} \delta_t = 0$, then AESCB verifies the same (state-of-the-art) regret as ESCB up to a multiplicative constant. For $m$-sets, knapsack sets, and source destination paths, we choose $\varepsilon = 1$. For spanning trees, matroids, matchings, and matroid intersection, we choose $\varepsilon = {1 \over 2}$ (see Section~\ref{sec:polynomial_time}). This choice of parameters does not require any knowledge about the time horizon $T$, nor about the unknown problem parameters $\theta$, nor about the minimal gap $\Delta_{\min}$. Nevertheless, if $\Delta_{\min}$ is known as well, we can select $\delta_t$ to yield an even better algorithm; however, knowing this parameter is by no means required.

        We will show that, with this choice of parameters, AESCB can be implemented in polynomial time (see Section~\ref{sec:polynomial_time}). A sketch of proof for Theorem~\ref{th:aescb_regret} is presented in the next subsection to further highlight the algorithm rationale, while the complete proof is presented in appendix.

    	\begin{theorem}[Regret of AESCB] \label{th:aescb_regret}
    		The regret of AESCB with parameters $(\varepsilon_t,\delta_t)$ admits the following upper bound for all $T \ge 1$:
    		\begin{align*}
    			R\!\left(T\right) &\leq C_4(m) + \frac{2\,d\,m^{3}}{\Delta_{\min}^{2}}
                                  + \frac{24  d \,f\left(T\right)}{(\min_{t \le T} \varepsilon_t)^2 \Delta_{\min}} \left\lceil \frac{\ln m}{1.61}\right\rceil ^{2} + 4 \sum_{t=1}^T \delta_t  \indic( \Delta_{\min} \le 4 \delta_t ) .
    		\end{align*}
    		with $f(t) = \ln t+ 4 m \ln\ln t$ and $C_4(m)$ a positive number that solely depends on $m$.

    		By corollary, for $\varepsilon_t = \varepsilon$ and $\lim_{t \to \infty} \delta_t = 0$, we have:
   			\begin{align*}
    			R\!\left(T\right)=\mathcal{O}\!\left(d\,(\ln m)^2\,\frac{1}{\Delta_{\min}}\,\ln T\right)\quad\text{as}\quad T\to\infty.
    		\end{align*}
    		Similarly, with $\varepsilon_t = \varepsilon$ and $\delta_t < {1 \over 4} \Delta_{\min}$, we have, for all $T \ge 1$:
   			\begin{align*}
    			R\!\left(T\right) &\leq C_4(m) + \frac{2\,d\,m^{3}}{\Delta_{\min}^{2}}
                                  + \frac{24  d \,f\left(T\right)}{\varepsilon^2 \Delta_{\min}} \left\lceil \frac{\ln m}{1.61}\right\rceil ^{2}.
    		\end{align*}
    	\end{theorem}

    \subsection{Theorem~\ref{th:aescb_regret}: Sketch of Proof}
    	The regret analysis of AESCB involves upper bounding the reward gap of the decision chosen at time $t$, $\Delta_{x(t)}$, by considering three cases. Define the following events:
        \begin{align*}
        		\cA_t &= \Big\{ \exists x \in \cX: | (\theta - \hat\theta(t))^\top x | \ge  \sqrt{\sigma^2(t)^\top x} \Big\} \\
        		\cB_t &= \{ \Delta_{x(t)}  \le 4 \delta_t \}
        \end{align*}
    	If $\cA_t$ occurs, since $\theta \in [0,1]^d$ and $\indic^\top x^\star \le m$:
    	\begin{align*}
    		 \Delta_{x(t)} \le \theta^\top x^\star  \le m.
    	\end{align*}
    	If $\cB_t$ occurs, by definition:
    	\begin{align*}
    		\Delta_{\min} \le \Delta_{x(t)}  \le 4 \delta_t.
    	\end{align*}
    	If $\cC_t = \overline{\cA_t} \cup \overline{\cB_t}$  occurs, the index of the optimal decision is greater than the optimal reward $\theta^\top x^\star$:
    	\begin{align*}
    	   \theta^\top x^\star  &\le \hat \theta(t)^\top x^\star + \sqrt{\sigma^2(t)^\top x^\star } \\
								&\le \delta_t + \hat\theta(t)^\top x(t)  + {1 \over \varepsilon_t} \sqrt{\sigma^2(t)^\top x(t)} \\
								&\le \delta_t + \theta^\top x(t)  + {2 \over \varepsilon_t} \sqrt{\sigma^2(t)^\top x(t) } \\
								&\le {1 \over 4} \Delta_{x(t)} + \theta^\top x(t)  + {2 \over \varepsilon_t} \sqrt{ \sigma^2(t)^\top x(t)}.
        \end{align*}
    	where we used the definition of AESCB, and the fact that $\cA_t$ and $\cB_t$ do not occur. Therefore, if $\cC_t$ occurs:
        \begin{align*}
        	\Delta_{x(t)} \le  {8 \over 3 \varepsilon_t} \sqrt{\sigma^2(t)^\top x(t)} \le {4 \over \varepsilon_t} \sqrt{\sigma^2(t)^\top x(t)}.
        \end{align*}
        Putting it together, we get:
        \begin{align*}
        	\Delta_{x(t)} & \le\Delta_{x(t)}\Big(\indic(\cA_{t})+\indic(\cB_{t})+\indic(\cC_{t})\Big)\\
                          & \le m\indic(\cA_{t})+4\delta_{t}\indic(\Delta_{\min}\le 4\delta_{t})+\Delta_{x(t)}\indic\left\{ \Delta_{x(t)}\le\frac{4}{\varepsilon_t}\sqrt{x(t)^{\top}\sigma^{2}(t)}\right\}.
        \end{align*}
        Taking expectations and summing over $t$:
        \begin{align*}
        	R(T) & =\sum_{t=1}^{T}\EE(\Delta_{x(t)})\\
                 & \le\sum_{t=1}^{T}m\PP(\cA_{t})+\sum_{t=1}^{T}4\delta_{t}\indic(\Delta_{\min}\le 4\delta_{t})+\sum_{t=1}^{T}\EE\Big(\Delta_{x(t)}\indic\left\{ \Delta_{x(t)}\le\frac{4}{\varepsilon_t}\sqrt{x(t)^{\top}\sigma^{2}(t)}\right\} \Big).
        \end{align*}
        The first term is bounded by a constant, since, using a concentration inequality, we may show that $\cA_t$ occurs with small probability. The last term can be bounded using (rather intricate) counting arguments as in the analysis of ESCB. The complete proof is presented in appendix.

\section{AESCB in Polynomial Time}\label{sec:polynomial_time}

   	We now show a technique to implement AESCB that ensures polynomial time complexity. While our methodology is generic, the precise value of the computational complexity depends on the combinatorial set $\cX$, and will be explained in details in Sections~\ref{subsec:msets} $-$ \ref{subsec:matchings}. Our approach involves three steps: rounding and scaling to ensure that the weights are integer, then solving the budgeted linear maximization $(P_3)$ several times, and finally maximizing over the budget to obtain the result. Given time $t$, statistics $\hat\theta(t)$ and $\sigma^2(t)$, and approximation factors $(\varepsilon_t,\delta_t)$, the method works as follows.

    \paragraph{Step 1: rounding and scaling.}
        Define $a(t)$ and $b(t)$:
        \begin{align*}
            \xi(t) &= \lceil {m / \delta_t} \rceil.\\
            a_i(t) &=  \lceil{ \xi(t) \hat\theta_i(n) } \rceil  \;,\; i \in \{1,...,d\}\\
            b_i(t) &= \xi(t)^2 \sigma_i^2(t) \;,\; i \in \{1,...,d\}
        \end{align*}
    \paragraph{Step 2: budgeted linear maximization.}
        For all $s \in \{0,...,m \xi(t)\}$, compute $\bar{x}^s(t)$, an $\varepsilon_t$-optimal solution to budgeted linear maximization problem $(P_3)$:
        \begin{align*}
            \bar{x}^s(t) \ge \varepsilon_t  \Big(\max_{x \in \cX: a(t)^\top x \ge s} \{ b(t)^\top x\} \Big)\quad\text{ and }\quad a(t)^\top \bar{x}^s(t) \ge s.
        \end{align*}

    \paragraph{Step 3: optimizing over a budget.}
        Return decision $x(t)$:
        \begin{align*}
            x(t) &= \bar{x}^{s^\star(t)}(t) \text{   with   }\\
            s^\star(t) &\in \mathop{\arg\max}_{s=0,...,m \xi(t)}\left\{ s + {1 \over \varepsilon_t} \sqrt{ b(t)^\top \bar{x}^{s}(t)} \right\}.
        \end{align*}

    $a(t)$ is defined using a ceiling operation in order to ensure that $a(t)^\top x$ has an integer value for any $x \in \cX$, while $b(t)$ does not need to have integer entries. Theorem~\ref{th:aescbimplementation} (see proof in appendix) states that this technique returns the decision chosen by AESCB, in a time proportional to solving the optimization problem $(P_3)$ at most $m \xi(t)$ times (where $\xi(t)$ is bounded by a polynomial in $d$), and that the input parameters $a(t)$ and $b(t)$ of $(P_3)$ are positive vectors and where the entries of $a(t)$ are in $\{1,...,\xi(t)\}$. In the next subsections, we show that, for many combinatorial sets of interest, one can solve $(P_3)$ with a time complexity bounded by a polynomial in the dimension, which implies that AESCB is indeed implementable in polynomial time.

    \begin{theorem}\label{th:aescbimplementation}
       	The above algorithm returns a decision $x(t) \in \cX$ verifying the AESCB definition:
       	\begin{align*}
       		\arg\max_{x \in \cX}  \{ \hat\theta(t)^\top x + \sqrt{\sigma^2(t)^\top x}   \}
       		\le \delta_t + \hat\theta(t)^\top x(t) + {1 \over \varepsilon_t} \sqrt{ \sigma^2(t)^\top x}
       	\end{align*}
       	It does so by solving optimization problem $(P_3)$ at most $m \xi(t)$ times with input parameters $a(t)$ and $b(t)$, where $a(t) \in \{1,...,\xi(t)\}^d$ and $b(t) \in \RR^d$.
    \end{theorem}

    Based on Theorem~\ref{th:aescbimplementation}, we now highlight how to find $\varepsilon$-optimal solutions to the optimization $(P_3)$. The complexity at each time step of AESCB with parameters $(\varepsilon_t,\delta_t)$ is summarized in Table~\ref{complexity_table}, where $\xi(t) = \lceil m \delta_t^{-1} \rceil$.

    Since $\delta_t$ can be chosen as a function that vanishes arbitrarily slowly, the dependency of the complexity on $t$ in Table~\ref{complexity_table} can be made arbitrarily mild. Additionally, if $\Delta_{\min}$ is known to the decision maker, one can set $\delta_t = {1 \over 4} \Delta_{\min}$, so that the computational complexity does not even depend on $t$ altogether. We emphasize once again that knowing $\Delta_{\min}$ is by no means necessary.

    We now consider time $t$ fixed and we drop the time index to simplify notation. We consider input parameters $a$ and $b$ for $(P_3)$ with $a \in \{1,...,\xi\}$. We do so for each type of combinatorial set in Sections~\ref{subsec:msets} $-$ \ref{subsec:matchings}. We provide a description of the algorithm, with pseudocode given in appendix.

    \subsection{$m$-sets}\label{subsec:msets}
        \begin{claim}
            Optimization problem $(P_3)$ with $\cX$ the set of $m$-sets can be solved exactly (i.e. $\varepsilon = 1$) in time $\cO( m^2 d \xi)$ using the algorithm below.
        \end{claim}
        In fact, since $m$-sets are a particular cases of a knapsack set with matrix $A = (1,...,1)^\top$, $k = 1$ and $c = (m)$, we can simply apply the algorithm for knapsack sets explained in Section~\ref{subsec:knapsack} below.

    \subsection{Knapsack sets}\label{subsec:knapsack}
        \begin{claim}
            Optimization problem $(P_3)$ with $\cX$ a knapsack set can be solved exactly (i.e. $\varepsilon = 1$) in time $\cO((\prod_{\ell=1}^k c_\ell) d m \xi)$ using the algorithm below.
        \end{claim}

        For $i \in \{0,...,d\}$, define the optimization problem
        \begin{align*}
            \underset{x}{\max} & b^\top x   \\
            \text{ subject to } & A x \le c \;,\; a^\top x \ge s \;,\; \\&  \sum_{j=1}^i x_j = 0 \;,\; x \in \{0,1\}^{d} \quad \quad (P_4(s,c,i))
        \end{align*}
        where we recall that $c \in \NN^k$ is a vector with integer entries. We denote by $\cV_4(s,c,i)$ the optimal value of this optimization program. Since $(P_4(s,c,0))$ reduces to $(P_3)$, it is sufficient to solve $(P_4(s,c,i))$ for $i \in \{0,...,d\}$. We do so using dynamic programming. Let $x^\star$ an optimal solution to $(P_4(s,c,i))$. If $x_{i+1}^\star = 1$, then
        \begin{align*}
            A (x^\star - e_i) = A x^\star - A e_i \le c - A e_i
        \end{align*}
        hence $x^\star - e_i$ is an optimal solution to $(P_4(\max{s-a_i,0},c - A e_i,i+1))$. If $x_i^\star = 0$, then $x^\star$ is an optimal solution to $(P_4(s,c,i+1))$. Therefore,
        \begin{align*}
           	\cV_4(s,c,i) = \max \{\cV_4(s,c,i+1),
          				  a_i + \cV_4(\max(s-a_i,0),c - A e_i,i+1) \}
        \end{align*}
        By recursion over $i$, $c$, and $s$, we can compute the value $\cV_4(s,c',i)$ for $s \in \{0,...,m \xi\}$, $c' \in \NN^k$ with $c' \le c$, and $i \in \{0,...,d\}$ in time $\cO((\prod_{\ell=1}^k c_\ell) d m \xi)$. The solution to $(P_3)$, denoted by $x^\star$, is then:
        \begin{align*}
            x^\star_i = \begin{cases}
                   			0 & \text{ if } \cV_4(s,c,i)  = \cV_4(s,c,i+1) \\
                   			1 & \text{ otherwise.}
               			\end{cases}
        \end{align*}
        Therefore, we can solve $(P_3)$ for all $s \le m \xi$ in time $\cO((\prod_{\ell=1}^k c_\ell) d m \xi)$. In particular, for $m$-sets, we can do so with a time complexity of $\cO(d m^2 \xi)$

    \subsection{Source destination paths}\label{subsec:paths}
        \begin{claim}
            Optimization problem $(P_3)$ with $\cX$ the set of paths between source $u$ and destination $v$ in $G = (V,E)$ a directed acyclic graph can be solved exactly (i.e. $\varepsilon = 1$) in time $\cO( m \xi |E| + |V| \ln |V| )$  using algorithm below.
        \end{claim}

       	Consider $v$ fixed throughout, and denote by $(P_3(u,s))$ this optimization problem and $\cV_3(u,s)$ its optimal value. If $s \le 0$, $(P_3(u,s))$ is simply the problem of finding the path from $u$ to $v$ maximizing $b^\top x$, since  $a$ has positive entries so that $a^\top x \ge 0 \ge s$, for all $x \in \cX$. Hence, we can compute $(P_3(u,s))$ for all $u$ with the Bellman-Ford algorithm in time $\cO(|V| |E|)$. 
       	
       	If $s > 0$, let $x^\star$ an optimal solution to $(P_3(u,s))$. Since $x^\star$ is a path from $u$ to $v$, there exists a unique $w \in V$ such that $x_{(u,w)}^{\star} = 1$, and  $(x^{\star} - e_{(u,w)})$ is a path from $w$ to $v$. In turn, we must have that $(x^{\star} - e_{(u,w)})$ is an optimal solution to $(P_3(w,\max(s-a_{(u,w)},0)))$. Therefore, we have the following dynamic programming equation:
        \begin{align*}
           	\cV_3(u,s) = \hspace{-0.3cm} \max_{w:(u,w) \in E} \hspace{-0.1cm} \{b_{(u,w)}  + \cV_3(w,\max\{s - a_{(u,w)},0\}) \}
        \end{align*}
       	As $a \in \{1,...,\xi\}^d$, if $\cV_3(u,s')$ is known for all $u \in V$ and all $s' \in \{0,...,s-1\}$, then applying the above relationship enables us to compute $\cV_3(u,s)$ for all $u \in V$. By recursion, we can compute $\cV_3(u,s)$ for all $s \in \{0,...,m \xi\}$ in time $\cO( (m \xi + |V|)|E|)$. The solution to $(P_3)$, denoted by $x^\star$, can also be computed by recursion. Using the same dynamic programming principle, denote by $x^\star(u,s)$ the solution of $(P_3(u,s))$, we have:
        \begin{align*}
            x^\star(u,s) &= e_{(u,w^\star(u,s))} + x^\star(w^\star(u,s),s - a_{(u,w^\star(u,s))}) \text{ with } \\
            w^\star(u,s) &\in \mathop{\arg\max}_{w:(u,w) \in E} \hspace{-0.05cm} \{b_{(u,w)}  + \cV_3(w,\max(s - a_{(u,w)},0)) \}
        \end{align*}
        By recursion, we can compute the solution to $(P_3)$ for all $s \in \{0,...,m \xi\}$ in time $\cO(  (m \xi+|V|) |E| )$.

    \subsection{Spanning Trees and Matroids}\label{subsec:trees}
        \begin{claim}
            Optimization problem $(P_3)$ with $\cX$ the set of spanning trees of graph $G = (V,E)$ can be solved with approximation ratio $\varepsilon = {1 \over 2}$ in time $\cO(|E| (\ln |V|)^2 + |V| (\ln |E|)^3)$ using the algorithm below. The same holds for any matroid.
        \end{claim}
        The algorithm is made of four steps, and is similar to that of~\cite{ravi1996} (see this reference for further details).

        \paragraph{Step 1: Lagrangian relaxation.}
            De
            \begin{align*}
                \underset{x}{\max} \{ b^\top x + \lambda(a^\top x - s)\} \text{ subject to } x \in X
            \end{align*}
            Denote by $M(\lambda)$ its value and $\cL(\lambda)$ the set of optimal solutions. Define $\lambda^\star = \arg\min_{\lambda \ge 0} M(z)$. Computing $M(\lambda)$ can be done in polynomial time using a greedy algorithm, since it is equivalent to maximizing a linear function over a matroid \cite{oxley2006}. Furthermore, $\lambda^\star$ can be found using Meggido's search technique~\cite{meggido1981}, as it involves minimizing a piecewise linear function.

        \paragraph{Step 2: candidate solutions.}
            For an arbitrarily small $\varepsilon > 0$, if $|\lambda - \lambda^\star| < \varepsilon$, we must have that $\cL(\lambda) \subset \cL(\lambda^\star)$. Therefore, by solving the Lagrangian relaxation of the problem for $\lambda=\lambda^\star +\varepsilon$ and $\lambda = \lambda^\star - \varepsilon$, we obtain two solutions $x^{+}$ and $x^{-}$ in $\cL(\lambda^\star)$ with $a^\top x^{+} \ge s$ and $a^\top x^{-} \le s$.

        \paragraph{Step 3: solution refining.}
            We now use an iterative procedure in order to find a good solution using candidates $x^+$ and $x^{-}$. Consider $e,e' \in E$ such that $x^+_{e} = x^-_{e'} = 1$ and $x^+_{e'} = x^-_{e} = 0$ and define $x = x^+ \\ \{e_e\} \cup \{e_{e'}\}$. If $a^\top x \ge s$, then replace $x^{+}$ by $x$ and otherwise replace $x^{-}$ by $x$ and repeat this procedure until $x^{+}$ and $x^{-}$ differ by exactly one element. Finally, return $x^+$. At each step of this procedure:
           	\begin{align*}
           		(b + \lambda a)^\top x  = (b + \lambda a)^\top x^+  = (b + \lambda a)^\top x^-
           	\end{align*}
           	therefore $x \in \cL(\lambda^\star)$. Denote by $x^\star$ the solution of $(P_3)$.

           	Since $x^+$ and $x^-$ are in $\cL(\lambda^\star)$:
           	\begin{align*}
           		b^\top x^{-} + \lambda( a^\top x^{-} - s) \ge b^\top x^{\star} + \lambda(  a^\top x^{\star} - s)
           	\end{align*}
           	Since $a^\top x^- \le s \le a^\top x^{\star}$, we deduce that $b^\top x^{-} \ge b^\top x^{\star}$. Since $x^{+}$ and $x^{-}$ differ by at most one element:
           	\begin{align*}
           		b^\top x^+ \ge b^\top x^{-} - \max_{e \in E} b_e \ge b^\top x^{\star} - \max_{e \in E} b_e
           	\end{align*}
           	So, after steps 1-3, we get $x$ such that $a^\top x \ge s$ and $b^\top x \ge b^\top x^{\star} - \max_{e \in E} b_e$.

        \paragraph{Step 4: a ${1 \over 2}$ optimal solution.}
            Lastly, we search over the two edges with largest weight to obtain a constant multiplicative approximation factor. For all sets of two edges $E'' \subset E$, $|E''| = 2$,  define $G' = (V,E')$ where
           	\begin{align*}
           		E' = \{e \in E \setminus E'': b_{e} \le \min_{e' \in E''} \{ b_{e''} \} \}
           	\end{align*}
           	and apply steps $1$ to $3$ to solve the problem where $G$ and $s$ are replaced by $G'$ and $s' = s - \sum_{e'' \in E''} a_{e''}$, where $x'(E'')$ is the solution found by steps 1-3. Finally, return $x(E'') = x'(E'')+ \sum_{e'' \in E''} e_{e''}$, for the value of $E''$ maximizing $a^\top x(E'')$. This yields a $1/2$ optimal solution in time $\cO(|E|^2 (\ln |V|)^2 + |V| (\ln |E|)^3)$ by the same arguments as that used in~\cite{ravi1996,berger2011}.

    \subsection{Matchings and Matroid Intersection}\label{subsec:matchings}
        \begin{claim}
            Optimization problem $(P_3)$ with $\cX$ the set of matchings of a bipartite graph $G = (V,E)$ can be solved with approximation ratio $\varepsilon = {1 \over 2}$ in time $\cO(|V|^3|E|^4)$  using the algorithm below. The same holds for any intersection of two matroids.
        \end{claim}

        The algorithm is made of four steps and is very similar of that of~\cite{berger2011}, which itself is inspired by the algorithm for matroids of~\cite{ravi1996}.

        \paragraph{Step 1: Lagrangian relaxation.}
            Define the Lagrangian relaxation of the problem:
            \begin{align*}
                \underset{x}{\max} \{ b^\top x + \lambda(a^\top x - s)\} \text{ subject to } x \in X
            \end{align*}
            denote by $M(\lambda)$ its value and $\cL(\lambda)$ the set of optimal solutions. Define $\lambda^\star = \arg\min_{\lambda \ge 0} M(\lambda)$. Computing $M(\lambda)$ can be done in time using the Hungarian algorithm, since it is maximizing a linear function over the set of matchings of a bipartite graph. Furthermore, $\lambda^\star$ can be found using Meggido's parametric search technique \cite{meggido1981} as it involves minimizing a piecewise linear function.

        \paragraph{Step 2: candidate solutions.}
            For an arbitrarily small $\varepsilon > 0$, if $|\lambda - \lambda^\star| < \varepsilon$ we must have that $\cL(\lambda) \subset \cL(\lambda^\star)$. So, by solving the Lagrangian relaxation of the problem for $\lambda=z^\star +\varepsilon$ and $\lambda = \lambda^\star - \varepsilon$, we obtain two solutions $x^{+}$ and $x^{-}$ in $\cL(\lambda^\star)$ with $a^\top x^{+} \ge s$ and $a^\top x^{-} \le s$.

        \paragraph{Step 3: solutions refining.}
            We now use an iterative procedure in order to find a good solution using  candidates $x^+$ and $x^{-}$. Define their symmetric difference $x' = x^+ \oplus x^-$. $x'$ is made of a disjoint union of paths and cycles. Take $x''$ as one of such paths or cycles, and define $x = x^- \oplus x''$. If $a^\top x \ge s$, then replace $x^{+}$ by $x$ and otherwise replace $x^{+}$ by $x$. Repeat this procedure until $x^{+}$ and $x^{-}$ differ by at most two elements (the symmetric difference $x^+ \oplus x^-$ decreases at each step). Finally, return $x^+$. At each step of this procedure:
           	\begin{align*}
           		(b + \lambda^\star a)^\top x  = (b + \lambda^\star a)^\top x^+  = (b + \lambda^\star a)^\top x^-
           	\end{align*}
           	therefore $x \in \cL(\lambda^\star)$. Denote by $x^\star$ the solution of $(P_3)$.

           	Since $x^+$ and $x^-$ are in $\cL(\lambda^\star)$:
           	\begin{align*}
           		b^\top x^{-} + \lambda^\star ( a^\top x^{-} - s) \ge b^\top x^{\star} + \lambda^\star(  a^\top x^{\star} - s)
           	\end{align*}
           	Since $a^\top x^- \le s \le a^\top x^{\star}$ we deduce that $b^\top x^{-}  \ge b^\top x^{\star}$. Since $x^{+}$ and $x^{-}$ differ by at most two elements:
           	\begin{align*}
           		b^\top x^+ \ge b^\top x^{-} - 2 \max_{e \in E} b_e \ge b^\top x^{\star} - 2 \max_{e \in E} b_e
           	\end{align*}
           	So, after steps 1-3, we get $x$ such that $a^\top x \ge s$ and $b^\top x \ge b^\top x^{\star} - 2 \max_{e \in E} b_e$.

        \paragraph{Step 4: an ${1 \over 2}$ optimal solution.}
            Lastly, we search over the four edges with largest weight to obtain a constant multiplicative approximation factor. For all sets of four edges $E'' \subset E$, $|E''| = 4$,  define $G' = (V,E')$ where
            \begin{align*}
           		E' = \{e \in e \in E \setminus E'': b_{e} \le \min_{e' \in E''} \{ b_{e''} \} \}
           	\end{align*}
           	and apply steps $1$ to $3$ to solve the problem where $G$ and $s$ are replaced by $G'$ and $s' = s - \sum_{e'' \in E''} a_{e''}$, where $x'(E'')$ is the solution found by steps 1-3. Finally, return $x(E'') = x'(E'')+ \sum_{e'' \in E''} e_{e''}$, for the value of $E''$ maximizing $a^\top x(E'')$. This yields a $1/2$ optimal solution in time $\cO(|V|^3 |E|^4)$ by the same arguments as that used in~\cite{berger2011}.

\section{Numerical Experiments}\label{sec:numerical}

We evaluate the performance of TS, CUCB, ESCB, and AESCB through numerical experiments. ESCB is implemented by casting the optimization problem $(P_2)$ as an ISOCP (Integer Second-Order Cone Programming) and using an ISOCP solver, see appendix for more details. As done in all prior work~\cite{cappe2012,combes2015}, we simulate ESCB and AESCB using $f(t) = \ln t$, neglecting the $(4 m \ln\ln t)$ term, which gives much better performance. This issue is discussed in~\cite{combes2015}. Our implementation of the four algorithms is available along with this article. For $m$-sets, we choose $m = \lfloor d/3 \rfloor$ and $\theta_i = 0.55$ for $i \le d/2$ and $\theta_i = 0.4$ for $i > d/2$. For source-destination paths, we consider the graph $G = (V,E)$, a complete directed acyclic graph, so that $(i,j) \in E$ if and only if $i < j$. The source is $1$, the destination is $|V|$ and $\theta_{(i,j)} = 0.4$ for $(i,j) \ne (1,|V|)$ and  $\theta_{(1,|V|)} = 0.55$. We have $d = |V|(|V|-1)/2$, $m =  |V| - 1$ and the optimal path is $\{(i, i + 1) for i\in\{1,2\dots |V|\}\}$. For spanning trees, we consider the graph $G =(V,E)$, a complete graph, and $\theta_{(i,j)} = 0.4$ for all $(i,j)$ with $i \ne 1$ and $\theta_{(1,j)} = 0.55$. We have $d = |V|(|V|-1)/2$, $m =  |V| - 1$; thus, the optimal decision is a star network. For matchings, we consider a complete bipartite graph $G = (V,E)$ with $V = V_1 \cup V_2$ and $|V_1| = |V_2|$,  $\theta_{(i,j)} = 0.4$ for all $(i,j)$, $i \ne j$ and $\theta_{(i,i)} = 0.55$. The optimal decision is $x_{(i,j)}^\star = \indic\{i = j\}$ and $d = |V_1| |V_2|$, $m = \min(|V_1|,|V_2|)$. In Tables~\ref{time_table} and~\ref{parameter_table} the results are presented in the format $\overline{z} \pm \Delta z$ where $\overline{z}$ is the empirical mean and $\Delta z$ is $1.96$ times the square root of the ratio between the empirical variance and the number of samples.

\begin{figure}%
        \centering
        \subfloat[a][$m$-sets, $d=10$]{\includegraphics[width=0.4\columnwidth]{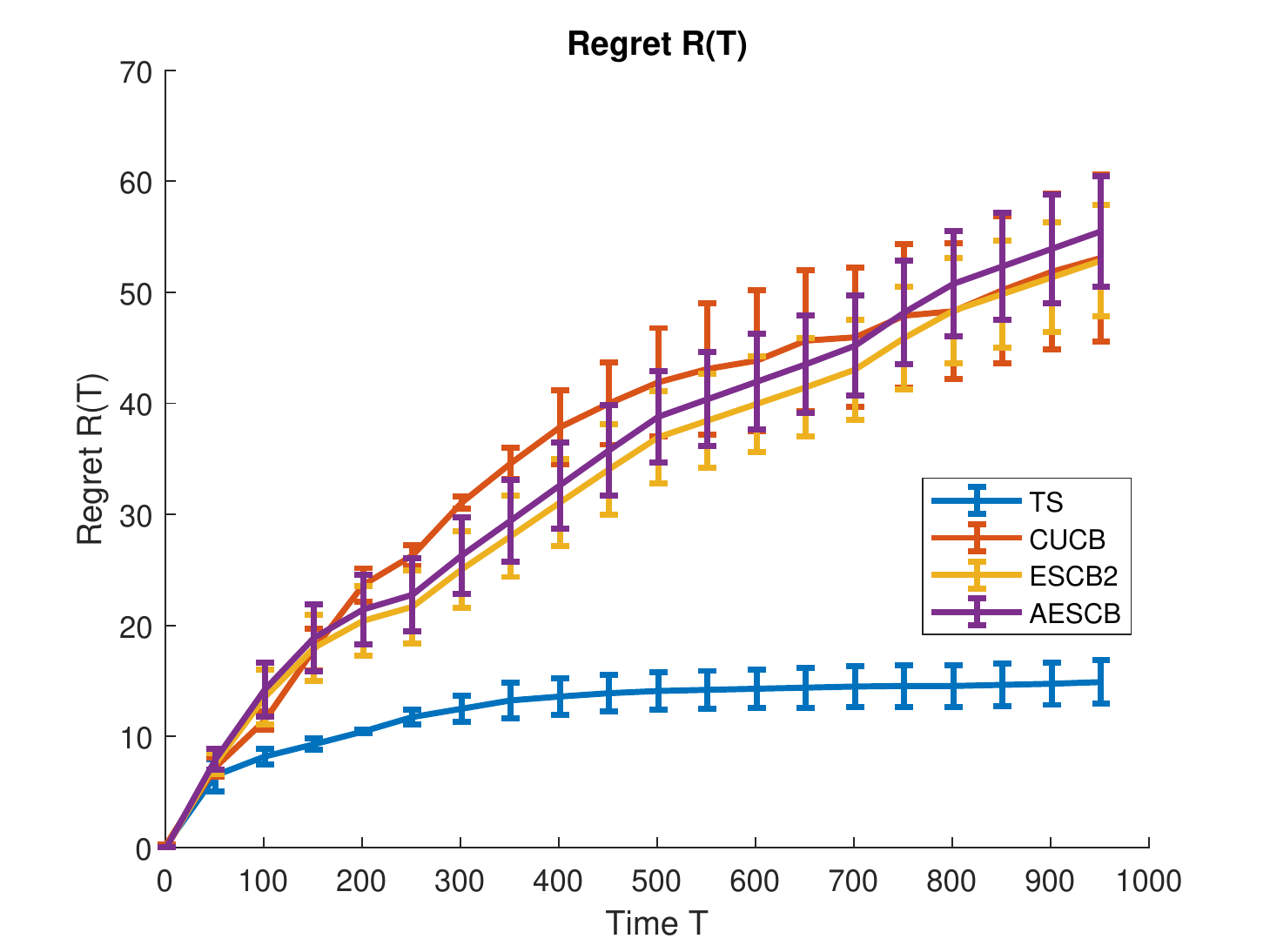}}
        \subfloat[b][spanning trees, $|V| = 5$]{\includegraphics[width=0.4\columnwidth]{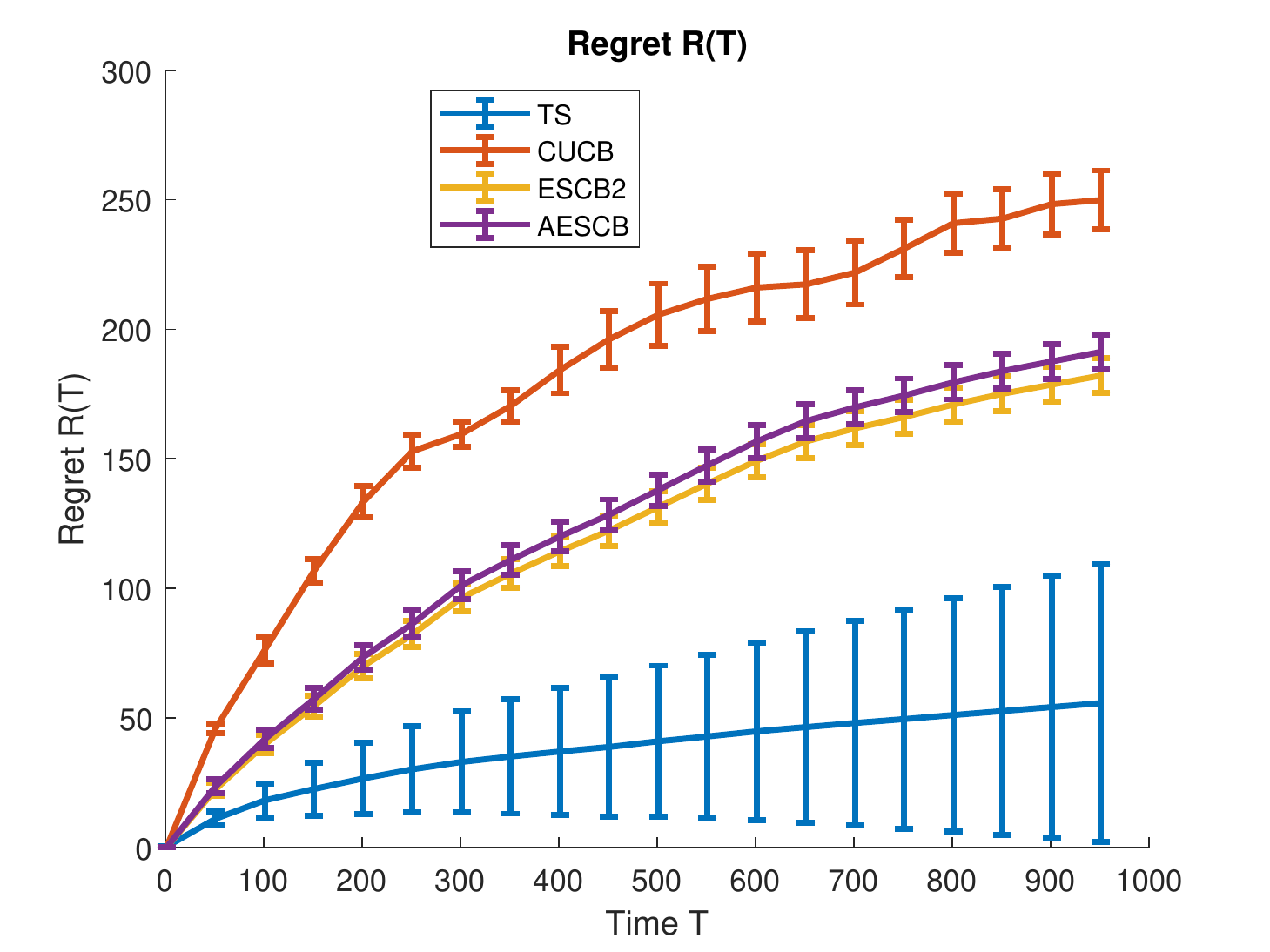}}%

        \subfloat[c][paths, $|V| = 10$]{\includegraphics[width=0.4\columnwidth]{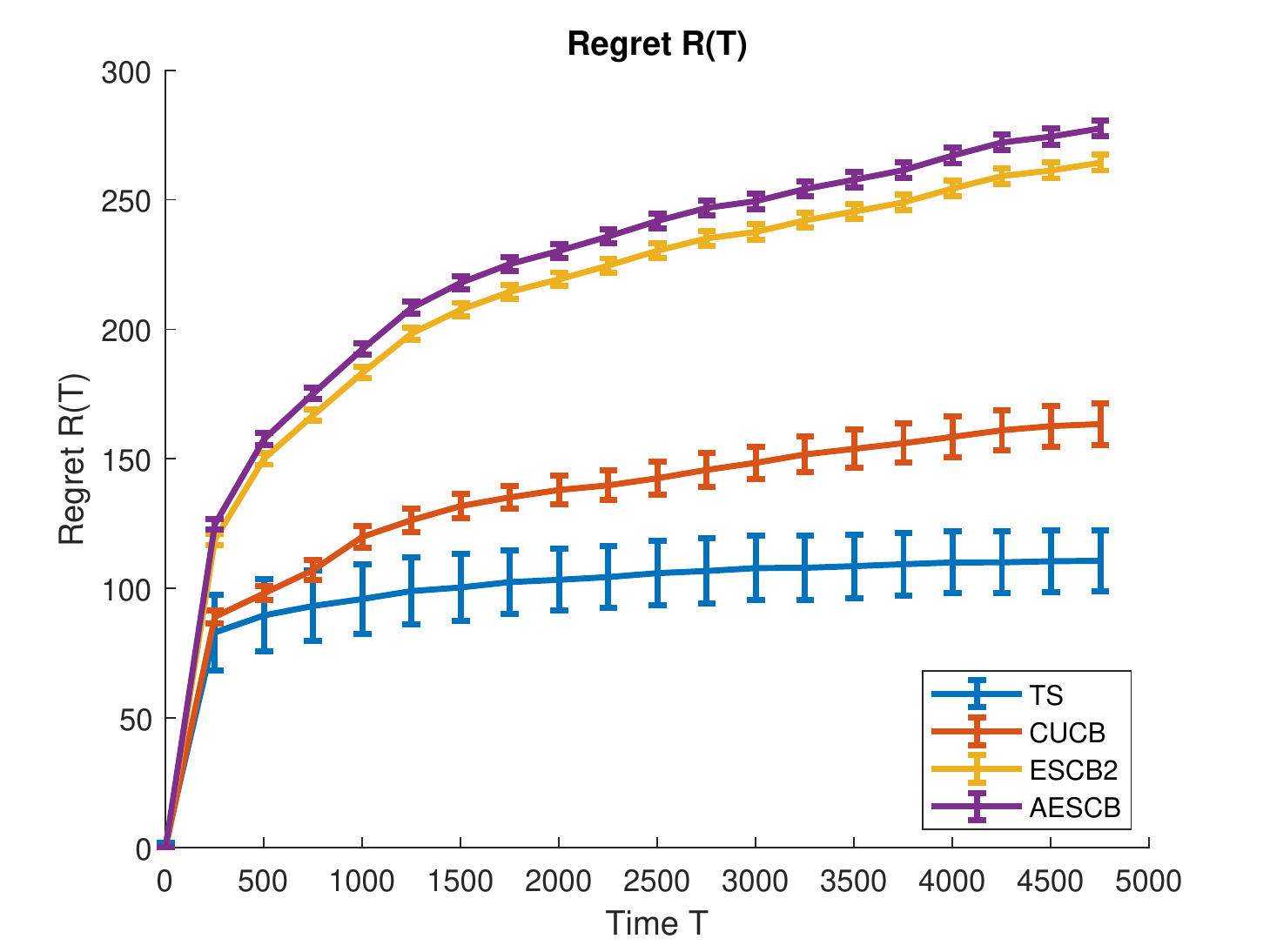}}
        \subfloat[d][matchings, $|V| = 10$]{\includegraphics[width=0.4\columnwidth]{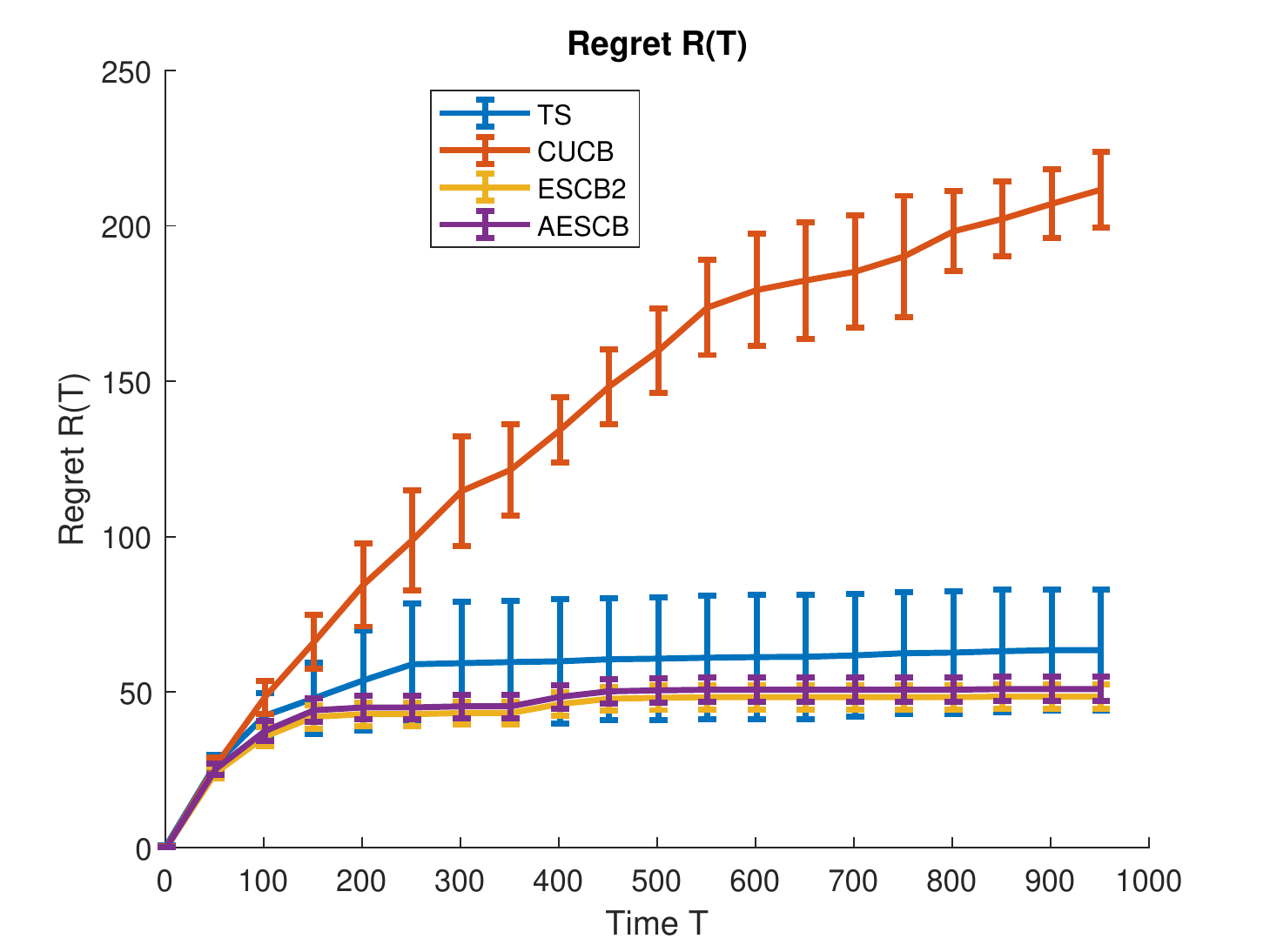}}
        \caption{Expected regret of algorithms}
        \label{fig_regret}
    \end{figure}

    \paragraph{Regret.}
        In Figure~\ref{fig_regret}, we present the expected regret of algorithms (with $95\%$ confidence intervals) averaged over 10 sample paths. We note that \textit{(i)} the regret of AESCB is very close to that of ESCB, so that the approximation comes at virtually no cost in terms of regret \textit{(ii)} ESCB and AESCB ouperform CUCB for matchings and spanning trees and CUCB performs better than ESCB and AESCB for paths \textit{(iii)} TS performs well on average; however, its regret has a lot of variability across sample paths, performing quite badly on some of them. Therefore, it is a ``risky'' algorithm to use, unlike the others.

    \paragraph{Tuning.} 
    In Table~\ref{parameter_table} we consider tuned versions of CUCB, ESCB and AESCB: 
    \begin{align*}
    &	x(t) \in \mathop{\arg\max}_{x \in \cX} \{  \hat{\theta}(t)^\top x +  \sum_{i=1}^d  x_i {\alpha \ln t \over  \sqrt{n_i(t)}} \} \hspace{5.5cm}\text{(tuned-CUCB)} \\
    	&x(t) \in \mathop{\arg\max}_{x \in \cX} \{ \hat\theta(t)^\top x + \sqrt{2 \alpha \sigma^2(t)^\top x} \} \hspace{5.5cm} \text{(tuned-ESCB)} \\
    &		\mathop{\arg\max}_{x \in \cX} \{ \hat\theta(t)^\top x + \sqrt{2 \alpha \sigma^2(t)^\top x}   \}
    		\le \delta_t + \hat\theta(t)^\top x(t) + {1 \over \varepsilon_t} \sqrt{2 \alpha \sigma^2(t)^\top x(t)}
\hspace{1cm} \text{(tuned-AESCB)}
   \end{align*} 
	where $\alpha \ge 0$ is a parameter, and $\alpha = {1 \over 2}$ corresponds to the normal versions of CUCB, ESCB and AESCB considered in Figure~\ref{fig_regret}. The regret obtained by choosing the best value of $\alpha$ is presented in bold. In general we notice that tuning $\alpha$ does yield some improvement in regret with respect to the standard algorithms (for $\alpha = {1 \over 2}$), however the improvement comes at the cost of careful tuning. We also notice that performing aggressive tuning by selecting $\alpha$ very small does not seem to yield good performance in general.

    \paragraph{Computation time.}
        In Table~\ref{time_table}, we present the computation times required to select an arm at time $t=1000$ for ESCB, AESCB, CUCB, and TS (with $95\%$ confidence intervals) as a function of the problem dimension $d$. We observe that the computation time for AESCB indeed appears to grow slowly in $d$, and that the computation times for all algorithms have the same magnitude.

	\begin{table}[h]
        \caption{Computing Time}
        \label{time_table}
        \vskip 0.15in
        \begin{center}
            \begin{small}
                \begin{sc}
                    \begin{tabular}{llccc}
                        \toprule
                        Problem   & Algo. & Time & [s] & \\
                        \midrule
                        $m$-sets  & ESCB      & $0.01 \pm 0.00$ & $0.02 \pm 0.00$ & $1.24 \pm 0.03$ \\
                                  & AESCB     & $0.00 \pm 0.00$ & $0.01 \pm 0.00$ & $0.10 \pm 0.10$ \\
                                  & CUCB      & $0.00 \pm 0.00$ & $0.00 \pm 0.00$ & $0.00 \pm 0.00$ \\
                                  & TS        & $0.00 \pm 0.00$ & $0.00 \pm 0.00$ & $0.00 \pm 0.00$ \\
                                  &           & $(d=10)$        & $(d=20)$        & $(d=50)$ \\
                        Paths     & ESCB      & $0.00 \pm 0.00$ & $0.02 \pm 0.00$ & $0.11 \pm 0.04$ \\
                                  & AESCB     & $0.00 \pm 0.00$ & $0.00 \pm 0.00$ & $0.05 \pm 0.00$ \\
                                  & CUCB      & $0.00 \pm 0.00$ & $0.00 \pm 0.00$ & $0.01 \pm 0.00$ \\
                                  & TS        & $0.00 \pm 0.00$ & $0.00 \pm 0.00$ & $0.01 \pm 0.00$ \\
                                  &           & $(d=10)$        & $(d=45)$        & $(d=190)$ \\
                        Trees     & ESCB      & $0.02 \pm 0.00$ & $0.06 \pm 0.05$ & $0.20 \pm 0.03$ \\
                                  & AESCB     & $0.00 \pm 0.00$ & $0.02 \pm 0.00$ & $0.04 \pm 0.01$ \\
                                  & CUCB      & $0.00 \pm 0.00$ & $0.00 \pm 0.00$ & $0.01 \pm 0.00$ \\
                                  & TS        & $0.00 \pm 0.00$ & $0.00 \pm 0.00$ & $0.01 \pm 0.00$ \\
                                  &           & $(d=10)$        & $(d = 45)$      & $(d=190)$ \\
                        Matchings & ESCB      & $0.01 \pm 0.00$ & $0.26 \pm 0.05$ & \\
                                  & AESCB     & $0.00 \pm 0.00$ & $0.18 \pm 0.01$ & \\
                                  & CUCB      & $0.00 \pm 0.00$ & $0.00 \pm 0.00$ & \\
                                  & TS        & $0.00 \pm 0.00$ & $0.00 \pm 0.00$ & \\
                                  &           & $(d=4)$         & $(d = 25)$      & \\
                        \bottomrule
                    \end{tabular}
                \end{sc}
            \end{small}
        \end{center}
        \vskip -0.1in
    \end{table}

 \begin{table}[h]
\caption{Regret with Parameter Tuning}
\label{parameter_table}
\vskip 0.15in
\begin{tabular}{llccccc}
\toprule
Problem & Algo. & \multicolumn{5}{c}{Regret}\tabularnewline
\midrule
 &  & $(\alpha=0.1)$ & $(\alpha=0.2)$ & $(\alpha=0.3)$ & $(\alpha=0.4)$ & $(\alpha=0.5)$\tabularnewline
$m$-sets  & ESCB & $31.6\pm24.8$ & $27.9\pm17.9$ & ${\bf 24.9\pm 14.3}$ & $25.1\pm13.5$ & ${\bf 24.9\pm13.0}$\tabularnewline
$(d=50)$ & AESCB & $36.1\pm25.7$ & $32.5\pm15.6$ & $29.6\pm15.1$ & $28.7\pm15.4$ & ${\bf 27.4\pm14.1}$\tabularnewline
 & CUCB & $164.8\pm129.9$ & $73.6\pm86.8$ & $27.5\pm10.3$ & ${\bf 16.6\pm5.1}$ & $24.8\pm13.9$\tabularnewline
Paths  & ESCB & $276.4\pm3.6$ & $275.0\pm4.5$ & $275.0\pm4.5$ & ${\bf 275.0\pm4.4}$ & ${\bf 275.0\pm4.4}$\tabularnewline
$(d=190)$ & AESCB & $284.0\pm3.8$ & $283.2\pm6.1$ & $283.2\pm7.1$ & ${\bf 282.3\pm4.6}$ & ${\bf 282.3\pm4.6}$\tabularnewline
 & CUCB & $95.9\pm43.0$ & $74.4\pm13.4$ & ${\bf 71.7\pm2.1}$ & $84.5\pm8.9$ & $102.2\pm5.7$\tabularnewline
Trees  & ESCB & $138.4\pm54.3$ & $137.1\pm51.6$ & ${\bf 128.1\pm37.9}$ & $148.1\pm33.3$ & $130.1\pm33.8$\tabularnewline
$(d=190)$ & AESCB & $166.0\pm57.9$ & $166.0\pm52.9$ & $155.2\pm38.9$ & $179.2\pm35.2$ & ${\bf 136.6\pm34.7}$\tabularnewline
 & CUCB & $544.1\pm30.6$ & $225.8\pm79.4$ & ${\bf 205.5\pm73.9}$ & $290.5\pm38.7$ & $327.8\pm47.2$\tabularnewline
Matchings  & ESCB & $108.9\pm119.4$ & $108.1\pm114.2$ & $108.1\pm114.4$ & ${\bf 325.1\pm107.9}$ & ${\bf 325.1\pm107.9}$\tabularnewline
 $(d=25)$ & AESCB & $360.2\pm123.3$ & $360.1\pm123.7$ & $357.9\pm117.4$ & $357.3\pm113.9$ & ${\bf 357.0\pm113.3}$\tabularnewline
 & CUCB & $838.2\pm423.4$ & ${\bf 365.4\pm104.2}$ & $431.1\pm60.6$ & $477.9\pm106.6$ & $566.9\pm76.7$\tabularnewline
\bottomrule
\end{tabular}
\vskip -0.1in
\end{table}

\section{Conclusion}\label{sec:conclusion}

We propose AESCB, the first algorithm which enjoys both the state-of-the art regret bound of ESCB and polynomial computational complexity. We believe our work opens two important research questions: $(i)$ Since TS has generally polynomial complexity and seems to work better than AESCB numerically, can one prove that it also has $R(T) = {\cal O}\Big( {d (\ln m)^2 (\ln T) \over \Delta_{\min} }\Big)$ regret in general? $(ii)$ What is the optimal trade-off between regret and computational complexity in combinatorial bandits?

\appendix

\section{On the NP-hardness of $P_2$} \label{P2nohard}
We discuss the $\mathcal{NP}$-hardness of $P_2$ by recalling some results of \cite{atamturk2017}. In fact the authors prove a more general result which is that $P_2$ is $\mathcal{NP}$-hard when the objective function $a^\top x + \sqrt{b^\top x}$ is replaced by $a^\top x + g(b^\top x)$ where $g$ is any strictly concave function.
\begin{proposition}(Proposition 1 of~\cite{atamturk2017})
	Consider $\cX = \{x \in \{0,1\}^d: \sum_{i=1}^d x_i = m \}$ and $g$ a stricly concave function. Then maximizing $x \mapsto a^\top x + g(b^\top x)$ over $\cX$ is $\mathcal{NP}$-hard.
\end{proposition}
\begin{proposition}(Proposition 3 of~\cite{atamturk2017})
	Consider $G = (V,E)$, a directed acyclic graph. Consider $\cX$ the set of paths in $G$ and $g$ a stricly concave function. Then maximizing $x \mapsto a^\top x + g(b^\top x)$ over $\cX$ is $\mathcal{NP}$-hard.
\end{proposition}
\begin{proposition}(Proposition 4 of~\cite{atamturk2017})
	Consider $G = (V,E)$, a complete bipartite graph. Consider $\cX$ the set of matchings in $G$ and $g$ a stricly concave function. Then maximizing $x \mapsto a^\top x + g(b^\top x)$ over $\cX$ is $\mathcal{NP}$-hard.
\end{proposition}

In summary, when $\cX$ is either the set of $m$-sets, the set of paths in a directed acyclic graph, or the set of matchings in a complete bipartite graph, then $P_2$ is $\mathcal{NP}$-hard.

\section{Regret of Thompson Sampling }
    The authors of \cite{wang2018} study a slightly more general problem than ours, and propose a regret bound for Thompson Sampling. Let us rephrase their bound with our notations. Their problem is a combinatorial semi bandit problem with a non linear reward function. Namely they consider decisions $x \in \cX \subset \{0,1\}^d$ in a combinatorial set, and the expected reward of decision $x \in \cX$ is given by a possibly non-linear function $r(\theta,x)$, where $\theta$ is a vector unknown to the learner. The expected reward function $r$ must  satisfy the Lipshitz condition:
    \begin{align*}
        |r(\theta,x) - r(\theta',x) | \le B | \theta'^\top x  - \theta^\top x |,   \forall \theta,\theta'
    \end{align*}
    with $B$ the Lipschitz constant. In our setting the reward function is $r(\theta,x) = \theta^\top x$ so that $B = 1$.

    Their main result is \cite{wang2018}[Theorem 1], which is the regret upper bound for Thompson Sampling:
    \begin{align*}
        R(T) \le C_5(\theta,\cX,\varepsilon) + 8 B^2 (\ln T) \sum_{i=1}^d  \max_{x \in \cX:x_i = 1, \Delta_x > 0} \left( { \indic^\top x  \over \Delta_x - 2 B( \indic^\top x^\star  + 2 ) \varepsilon} \right)   \,\,\,\, \forall \varepsilon > 0
    \end{align*}
    where $C_5(\theta,\cX,\varepsilon)$ is a positive number which does not depend on $T$. In our setting the reward function is $r(\theta,x) = \theta^\top x$ so that $B = 1$ and in the worse case, we will have for all $i$:
    \begin{align*}
        \max_{x \in \cX:x_i = 1, \Delta_x > 0} {\indic^\top x \over \Delta_x} = {m \over \Delta_{min}},
    \end{align*}
    Hence the upper bound for the regret of Thompson Sampling provided by \cite{wang2018}[Theorem 1] scales as:
    \begin{align*}
        R(T) = \cO \left\{ {d m \ln T \over \Delta_{\min}} \right\} \,,\, T \to \infty
    \end{align*}
    and this bound does not match the (smaller) regret upper bound of algorithms such as ESCB and AESCB which is
    \begin{align*}
        R(T) = \cO \left\{ {d (\ln m)^2 \ln T \over \Delta_{\min}} \right\} \,,\, T \to \infty
    \end{align*}

\section{Casting Optimization problem $(P_2)$ as a ISOCP}
    The optimization problem $(P_2)$, whose definition is:
    \begin{align*}
        \underset{x}{\text{maximize }} \{ a^\top x + \sqrt{b^\top x} \} \text{  subject to }  x \in \cX \quad (P_2)
    \end{align*}
    can be cast as an Integer Second-Order Cone Program (MISOCP), which enables one to solve it using a standard ISCOP solver. Indeed, the objective function features a geometric mean, which a special case of hyperbolic constraint~\cite{lobo1998}. Problem $(P_2)$ can be rewritten as:
    \begin{align*}
        \underset{(x,t)}{\text{maximize }} \{ a^\top x + t \} \text{  subject to }  x \in \cX \qquad\text{and}\qquad t^2\leq b^\top x
    \end{align*}
    Applying the transformation proposed in \cite{lobo1998}[Section 2.3], the hyperbolic constraint can be written as a SOCP (Second-Order Cone Program):
    \begin{align*}
        \underset{(x,t)}{\text{maximize }} \{ a^\top x + t \} \text{  subject to }  x \in \cX \qquad\text{and}\qquad \left\|\begin{bmatrix}\begin{array}{l} 2t\\ b^\top x-1\\ \end{array}\end{bmatrix} \right\| \leq b^\top x+1
    \end{align*}
    Even though the constraints defining $\cX$ ensure that optimizing a linear objective over $\cX$ yields an integer solution, this is no more the case with the new formulation. Hence, integrality constraints must be added for the relevant variables.
    \begin{align*}
        \underset{(x,t)}{\text{maximize }} \{ a^\top x + t \} \text{  subject to }  x \in \cX \qquad\text{and}\qquad \left\|\begin{bmatrix}\begin{array}{l} 2t\\ b^\top x-1\\ \end{array}\end{bmatrix} \right\| \leq b^\top x+1\qquad\text{and}\qquad x\in\{0,1\}^d
    \end{align*}
    This formulation is an ISOCP and can be readily solved to optimality by existing software such as CPLEX, Gurobi, Pajarito~\cite{coey2018}, Mosek, or Xpress.

\section{Proof of Theorem 4.4}

    {\bf Theorem 4.4.} The regret of AESCB with parameters $(\varepsilon_t,\delta_t)$ admits the following upper bound for all $T \ge 1$:
            \begin{align*}
                R\!\left(T\right) &\leq C_4(m) + \frac{2\,d\,m^{3}}{\Delta_{\min}^{2}}
                                  + \frac{24  d \,f\left(T\right)}{(\min_{t \le T} \varepsilon_t)^2 \Delta_{\min}} \left\lceil \frac{\ln m}{1.61}\right\rceil ^{2} + 4 \sum_{t=1}^T \delta_t  \indic( \Delta_{\min} \le 4 \delta_t ) .
            \end{align*}
            with $f(t) = \ln t+ 4 m \ln\ln t$ and $C_4(m)$ a positive number that solely depends on $m$.

            By corollary, for $\varepsilon_t = \varepsilon$ and $\lim_{t \to \infty} \delta_t = 0$, we have:
               \begin{align*}
                R\!\left(T\right)=\mathcal{O}\!\left(d\,(\ln m)^2\,\frac{1}{\Delta_{\min}}\,\ln T\right)\quad\text{as}\quad T\to\infty.
            \end{align*}
            Similarly, with $\varepsilon_t = \varepsilon$ and $\delta_t < {1 \over 4} \Delta_{\min}$, we have, for all $T \ge 1$:
               \begin{align*}
                R\!\left(T\right) &\leq C_4(m) + \frac{2\,d\,m^{3}}{\Delta_{\min}^{2}}
                                  + \frac{24  d \,f\left(T\right)}{\varepsilon^2 \Delta_{\min}} \left\lceil \frac{\ln m}{1.61}\right\rceil ^{2}.
            \end{align*}

    \vspace{1cm}

    {\bf Proof.}
    We decompose the regret based on three events:
    \begin{itemize}
        \item $G_{t}$: the estimate $\hat{\theta}(t)$ deviates abnormally from $\theta$ so that:
        \begin{align*}
            G_{t}=\left\{ \theta^\top x^\star \ge \hat{\theta}(t)^\top x^\star + \sqrt{\sigma^2(t)^\top x^\star} \right\} .
        \end{align*}
        \item $H_{t}$: the reward of the decision chosen at time $t$ is poorly estimated, namely:
        \begin{align*}
            H_{i,t}=\left\{ x_{i}\!\left(t\right)=1,\quad\left|\hat{\theta}_{i}\!\left(t\right)-\theta_{i}\right|\geq\frac{\Delta_{\min}}{2\,m}\right\} \text{ and } H_{t}=\bigcup_{i=1}^{d}H_{i,t}.
        \end{align*}
        \item $I_t$: the reward gap of the decision chosen at time $t$ is small
        \begin{align*}
            I_t = \left\{ \Delta_{x(t)} \le 4 \delta_t \right\}
        \end{align*}
    \end{itemize}
    Of course, most of the time, $\overline{G_{t}}$ and $\overline{H_{t}}$ occur, since both $G_{t}$ and $H_{t}$ have small probability. Therefore, $G_{t}$ and $H_{t}$ cause only a constant regret as we shall see. Also, $I_t$ causes a regret that is at most $4 \delta_t $. For all $x \in \cX$ and $t \ge 1$, we define the exploration bonus $E_t(x)$ of decision $x$ at time $t$:
    \begin{align*}
        E_t(x)= \sqrt{\sigma^2(t)^\top x}.
    \end{align*}

    \paragraph{Generic regret bound.}

        Recall that the regret is defined as:
        \begin{align*}
            R\!\left(T\right)=\mathbb{E}\left\{ \sum_{t=1}^{T}\Delta_{x\left(t\right)}\right\} = \mathbb{E}\left\{ \sum_{t=1}^{T}\Delta_{x\left(t\right)} \indic \{x(t) \ne x^\star  \}\right\},
        \end{align*}
        Decomposing according to the occurrence of $G_t$ and $H_t$, we get:
        \begin{align*}
            R\!\left(T\right) & \leq \mathbb{E}\left\{ \sum_{t=1}^{T}\indic\!\left\{ G_{t} \right\} \,\Delta_{x\left(t\right)}\right\} +\mathbb{E}\left\{ \sum_{t=1}^{T}\indic\!\left\{ H_{t} \right\} \,\Delta_{x\left(t\right)}\right\}
            + \mathbb{E}\left\{ \sum_{t=1}^{T}\indic\!\left\{ I_{t} \right\} \,\Delta_{x\left(t\right)}\right\}
            \\ & \;\;\; +\mathbb{E}\left\{ \sum_{t=1}^{T}\indic\!\left\{ \overline{G_{t}},\overline{H_{t}},\overline{I_{t}}, x\!\left(t\right)\neq x^\star\right\} \,\Delta_{x\left(t\right)}\right\} .
        \end{align*}

        Define $\overline{\varepsilon_T} = \min_{t \le T} \varepsilon_t$. The last term can be rewritten in terms of the following event:
        \begin{align*}
            F_{t}=\left\{ \Delta_{x\left(t\right)}\leq {4 \over \overline{\varepsilon}_T} E_t(x(t)) \right\} ,
        \end{align*}
        Let us prove that $\left(\overline{G_{t}}\cap\overline{H_{t}}\cap \overline{I_t} \cap  \{ x\!\left(t\right)\neq x^\star \}\right)\subset F_{t}$. Assume that $\left(\overline{G_{t}}\cap\overline{H_{t}}\cap  \overline{I_t} \cap\{ x\!\left(t\right)\neq x^\star \}\right)$ occurs. Then,
        \begin{align*}
            \theta^\top x^\star  &\le \hat \theta(t)^\top x^\star +  E_t(x^\star)\\
                                 &\le \max_{x \in \cX} \{ \hat \theta(t)^\top x +  E_t(x) \}\\
                                 &\le  \delta_t + \hat\theta(t)^\top x(t)  + {1 \over \varepsilon_t} E_t(x(t))\\
                                 &\le  \delta_t + \theta^\top x(t) + {\Delta_{x(t)} \over 2} + {1 \over \varepsilon_t} E_t(x(t)) \\
                                 &\le  \theta^\top x(t) + {3 \Delta_{x(t)} \over 4} + {1 \over \varepsilon_t} E_t(x(t))
        \end{align*}
           where we successively used the fact that $\overline{G_{t}}$ occurs, the definition of AESCB, the fact that $\hat\theta(t)^\top x(t) \le \theta^\top x(t) + {\Delta_{x(t)} \over 2}$ since $\overline{H_{t}}$ occurs, and the fact that $\delta_t \le {\Delta_{x(t)} \over 4}$ since $\overline{I_{t}}$ occurs. Therefore,
        \begin{align*}
            {\Delta_{x(t)} \over 4} \le {1 \over \varepsilon_t} E_t(x(t))  \le {1 \over \overline{\varepsilon}_T} E_t(x(t))
        \end{align*}
        so that $F_t$ indeed occurs.

        As a consequence, the regret is upper bounded by the sum of four terms:
        \begin{align*}
            \hspace{-0.5cm} R\!\left(T\right) & \leq \mathbb{E}\left\{ \sum_{t=1}^{T}\indic\!\left\{ G_{t} \right\} \,\Delta_{x\left(t\right)}\right\} +\mathbb{E}\left\{ \sum_{t=1}^{T}\indic\!\left\{ H_{t} \right\} \,\Delta_{x\left(t\right)}\right\} + \mathbb{E}\left\{ \sum_{t=1}^{T}\indic\!\left\{ I_{t} \right\} \,\Delta_{x\left(t\right)}\right\} + \mathbb{E}\left\{ \sum_{t=1}^{T}\indic\!\left\{ F_{t} \right\} \,\Delta_{x\left(t\right)}\right\}
        \end{align*}

    \paragraph{First term: poor reward estimation.}

        Recall that for any $x$ we have $\Delta_x \le \theta^\top x^\star \le m$, since $\theta \in [0,1]^d$ and $\max_{x \in \cX} \indic^\top x = m$. Therefore, by applying \cite{combes2015}[Theorem 3]:
        \begin{align*}
            \mathbb{E}\left\{ \sum_{t=1}^{T}\indic\!\left\{ G_{t} \right\} \,\Delta_{x\left(t\right)}\right\} \leq m \mathbb{E}\left\{ \sum_{t=1}^{T}\indic\!\left\{ G_{t} \right\} \right\} = m \sum_{t=1}^{\infty} \PP(G_t) \leq C_4(m).
        \end{align*}
        where $C_4(m)$ is a positive number which only depends on $m$, as stated by \cite{combes2015}[Theorem 3].

    \paragraph{Second term: poor choice of item.}

        We turn to the second term, using a union bound
        \begin{align*}
            \mathbb{P}\!\left(H_{t}\right) & =\mathbb{P}\!\left(\bigcup_{i=1}^{d}H_{t,i}\right)  \leq\sum_{i=1}^{d}\mathbb{P}\!\left(H_{t,i}\right).
        \end{align*}
        Using once again the fact that $\Delta_{x(t)} \le m$, the regret due to $H_{t} = \bigcup_{i=1}^{d}H_{t,i}$ is bounded by using a union bound:
        \begin{align*}
            \mathbb{E}\left\{ \sum_{t=1}^{T}\indic\!\left\{ H_{t}\right\} \,\Delta_{x\left(t\right)}\right\} \le m \mathbb{E}\left\{ \sum_{t=1}^{T}\indic\!\left\{ H_{t} \right\} \right\} = m \sum_{t=1}^{T} \PP( H_{t} ) \le m \sum_{t=1}^{T} \sum_{i=1}^d \PP( H_{t,i} ).
        \end{align*}
        By definition of $H_{t,i}$, we get
        \begin{align*}
            \mathbb{E}\left\{ \sum_{t=1}^{T}\indic\!\left\{ H_{t}\right\} \,\Delta_{x\left(t\right)}\right\} \leq m \sum_{t=1}^{T}\sum_{i=1}^{d} \,\mathbb{P}\!\left(x_i(t)=1,\quad\left|\hat{\theta}_{i}\!\left(t\right)-\theta_{i}\right|\geq\frac{\Delta_{x\left(t\right)}}{2\,m}\right)
        \end{align*}
        Using Hoeffding's inequality, this probability can be bounded by:
        \begin{align*}
            \mathbb{E}\left\{ \sum_{t=1}^{T}\indic\!\left\{ H_{t},x\!\left(t\right)\neq x^\star\right\} \,\Delta_{x\left(t\right)}\right\} \leq m \sum_{t=1}^{T}\sum_{i=1}^{d} e^{- t \left(\frac{\Delta_{\min}}{m}\right)^{2}} \le { m d \over 1 - e^{- \left(\frac{\Delta_{\min}}{m}\right)^{2}} } \le {m^3 d (1 + {\Delta_{\min}^2 \over m^2}) \over \Delta_{\min}^2} \\
            \le {2 m^3 d \over \Delta_{\min}^2}
        \end{align*}
        where we recognize a geometric series and use the elementary inequality $e^z \ge 1 + z$ for all $z$, which gives $e^{-z} \le (1+z)^{-1}$ and
        $(1 - e^{-z})^{-1}  \le (1 + z) z^{-1}$. We also used the fact that $\Delta_{\min} \le m$.

    \paragraph{Third term: small reward gap.} By definition, the third term is:
    \begin{align*}
        \mathbb{E}\left\{ \sum_{t=1}^{T}\indic\!\left\{ I_{t} \right\} \,\Delta_{x\left(t\right)}\right\} =  \mathbb{E}\left\{ \sum_{t=1}^{T}\indic\!\left\{ \Delta_{x(t)} \le 4 \delta_t \right\} \,\Delta_{x\left(t\right)}\right\}
        \le  4 \sum_{t=1}^{T} \delta_t \indic\!\left\{ \Delta_{\min} \le 4 \delta_t \right\}.
    \end{align*}

    \paragraph{Fourth term: dominant term.}

        We now consider the event $\Delta_{x\left(t\right)}\leq {4 \over \overline{\varepsilon}_T} E_{t}\!\left(x\!\left(t\right)\right)$. Squaring, we get:
        \begin{align*}
            \Delta_{x\left(t\right)}^2 \le {16 \over \overline{\varepsilon}_T^2}  E_{t}\!\left(x\!\left(t\right)\right)^2 = {16 \over \overline{\varepsilon}_T^2} \sigma^2(t)^\top x(t) = {8 f(t) \over \overline{\varepsilon}_T^2} \sum_{i=1}^d {x_i \over n_i(t)}.
        \end{align*}
        where we used the definition of $E_t$ and that of $\sigma^2(t)$. If this event happens, it means that there exists a subset of indices $i=1,...,d$ such that the number of samples $n_i(t)$ is small. We further decompose this event as follows.

        Consider $(\alpha_j)_{j \in \NN}$ and $(\beta_j)_{j \in \NN}$, two positive, non-increasing sequences verifying the following properties:
        \begin{itemize}
            \item $\lim_{j\to\infty}\alpha_{j} = \lim_{j\to\infty}\beta_{j} = 0$
            \item $\lim_{j\to+\infty}\frac{\beta_{j}}{\sqrt{\alpha_{j}}}=0$
            \item $\beta_0 = 1$
        \end{itemize}
        We can define $j_{0}$ as the first integer $j$ such that $\beta_{j}\leq\frac{1}{m}$ and define $l$ as the sum
        \begin{align*}
            l=\frac{\beta_{j_{0}}}{\alpha_{j_{0}}}+\sum_{j=1}^{j_{0}}\frac{\beta_{j-1}-\beta_{j}}{\alpha_{j-1}},
        \end{align*}
        Sequences $(\alpha_j)_{j \in \NN}$ and $(\beta_j)_{j \in \NN}$ are fixed, and their exact value will be specified later.

        For all $j\in\mathbb{N}$, define the following sets:
        \begin{align*}
           \hspace{-0.5cm} S_{t}^{j}&=\left\{ i \in \{ 1,...,d\}: x_i(t) = 1 , n_{i}\!\left(t\right)\leq\alpha_{j}\,\frac{2\,f\!\left(t\right)\, \, g(m) }{\Delta_{x\left(t\right)}^2}\right\} \text{ if }j\geq1 \text{ and  } S_{0}^{j}=\left\{ i \in \{ 1,...,d\} :  x_i(t) = 1 \right\}\text{ otherwise}
        \end{align*}
        where $g(m)$ is defined as $g(m) = {4 m l \over \overline{\varepsilon}_T}$ with $l$ a constant to be defined later.

        Since $j \mapsto \alpha_j$ is decreasing and $\lim_{j\to\infty}\alpha_{j}=0$, the sequence $S_{t}^{j}$ is decreasing for set inclusion; moreover, there is an index $j_{\emptyset}$ such that $S_{n}^{j_{\emptyset}}=\emptyset$:
        \begin{align*}
            \emptyset = S_{t}^{j_{\emptyset}} \subset S_{t}^{j_{\emptyset}-1} \subset ... \subset  S_{t}^{1} \subset S_{t}^{0}
        \end{align*}
        Define the event $A_{t}^{j}$ as:
        \begin{align*}
            A_{t}^{j}=\left\{ \left|S_{t}^{j}\right|\geq m\,\beta_{j}\qquad\text{and}\qquad\forall k<j,\quad\left|S_{t}^{j}\right|<m\,\beta_{k}\right\} .
        \end{align*}
        By assumption, we have:
        \begin{align*}
            \left|S_{t}^{0}\right|=m\,\beta_{0} = m.
        \end{align*}
        Finally, also define the events $A_{t}$ as the following unions:
        \begin{align*}
            A_{t}=\bigcup_{j=1}^{+\infty} A_{t}^{j}
        \end{align*}

        We have that $A_{t}$ is a finite union of events $ A_{t}^{j}$:
        \begin{align*}
            A_{t}=\bigcup_{j=1}^{j_{0}}A_{t}^{j}.
        \end{align*}
        Indeed, for all $j>j_{0}$, due to $\beta_{j_{0}}\leq1/m$ and the fact that $\beta_{j}$ is a decreasing sequence,
        \begin{align*}
            m\,\beta_{j}\leq m\,\frac{1}{m}=1.
        \end{align*}
        Thus, by definition of the event $A_{t}^{j}$, we have that:
        \begin{align*}
            A_{t}^{j}=\left\{ \left|S_{t}^{j}\right|\geq1\qquad\text{and}\qquad\forall k<j_{0},\quad\left|S_{t}^{j}\right|<m\,\beta_{j}\qquad\text{and}\qquad\forall k\in\left[j_{0},j-1\right],\quad\left|S_{t}^{j}\right|=0\right\} .
        \end{align*}
        However, the same set $S_{t}^{j}$ cannot be both empty and have at least one element. In other words, the event $A_{t}^{j}$ cannot happen for $j>j_{0}$.

        Under event $\overline{A}_{t}$, the sum $\sum_{i=1}^d {x_i(t) \over n_{i}(t)}$ can be bounded. The event $\overline{A}_{t}$ is, by De Morgan's law (recall that $j_{0}$ is finite):
        \begin{align*}
            \overline{A}_{t} & =\bigcap_{j=1}^{j_{0}}\overline{A}_{t}^{j}\\
                             & =\bigcap_{j=1}^{j_{0}}\left\{ \left|S_{t}^{j}\right|<m\,\beta_{j}\qquad\text{or}\qquad\exists k<j,\quad\left|S_{t}^{j}\right|\geq m\,\beta_{j}\right\} ,\qquad\text{by definition of }A_{t}^{j}\\
                             & =\bigcap_{j=1}^{j_{0}}\left[\left\{ \left|S_{t}^{j}\right|<m\,\beta_{j}\right\} \cup\left\{ \bigcup_{k=1}^{j-1}\left|S_{t}^{j}\right|\geq m\,\beta_{j}\right\} \right]\\
                             & =\bigcap_{j=1}^{j_{0}}\left\{ \left|S_{t}^{j}\right|<m\,\beta_{j}\right\} ,\qquad\text{as the latter events are not possible if the first holds}\\
                             & =\bigcap_{j=1}^{j_{0}-1}\left\{ \left|S_{t}^{j}\right|<m\,\beta_{j}\right\} \cap\left\{ \left|S_{t}^{j_{0}}\right|<m\,\beta_{j_{0}}\right\} .
        \end{align*}
        Since $\beta_{j_{0}}\leq1/m$, the last event can be written as $\left|S_{t}^{j_{0}}\right|< {m \over m} =1$. A set whose cardinality is strictly less than one must be empty, thus:
        \begin{align*}
            \overline{A}_{t}=\bigcap_{j=1}^{j_{0}-1}\left\{ \left|S_{t}^{j}\right|<m\,\beta_{j}\right\} \cap\left\{ \left|S_{t}^{j_{0}}\right|=0\right\} .
        \end{align*}
        If event $\overline{A}_{t}$ happens, then:
        \begin{align*}
            \overline{S}_{t}^{j} &=\left\{i=1,...,d: x_i(t) = 1 , i\not\in S_{t}^{j}\right\} =\left\{ i=1,...,d: x_i(t) = 1 , n_{i}\!\left(t\right)>\alpha_{j}\,\frac{2\,f\!\left(t\right)\,\,g(m)}{\Delta_{x\left(t\right)}^2}\right\} , \\
            \overline{S}_{t}^{j_{0}} & = \{i=1,...,d: x_i(t) = 1 \}.
        \end{align*}
        Indeed, due to the fact that $S_{t}^{j}$ is a decreasing sequence for set inclusion, the complement $\overline{S}_{t}^{j}$ must be an increasing sequence for set inclusion. This implies that:
        \begin{align*}
            \{i=1,...,d: x_i(t) = 1 \}=\bigcup_{j=1}^{j_{0}}\left(\overline{S}_{t}^{j}\backslash\overline{S}_{t}^{j-1}\right).
        \end{align*}
        Thus,
        \begin{align*}
            \sum_{i=1}^d {x_i(t) \over  n_i(t)} = \sum_{j=1}^{j_{0}} \sum_{i\in\overline{S}_{t}^{j}\backslash\overline{S}_{t}^{j-1}} {x_i(t) \over n_i(t)}.
        \end{align*}
        Using the definition of $\overline{S}_{t}^{j}$, one might write that, if $\overline{S}_{t}^{j}$ holds, then:
        \begin{align*}
            \sum_{i\in\overline{S}_{t}^{j}\backslash\overline{S}_{t}^{j-1}} {x_i(t) \over n_i(t)}  < \frac{\Delta_{x\left(t\right)}^2}{2\,f\!\left(t\right)\, g(m) \,\alpha_{j}} \sum_{i\in\overline{S}_{t}^{j}\backslash\overline{S}_{t}^{j-1}} x_i(t) = \frac{\Delta_{x\left(t\right)}^2}{2\,f\!\left(t\right)\, g(m) } { |\overline{S}_{t}^{j}\backslash\overline{S}_{t}^{j-1}| \over \alpha_j}.
        \end{align*}
        This implies that the previous sum is bounded by:
        \begin{align*}
            \sum_{i=1}^d {x_i(t) \over  n_i(t)} = \sum_{j=1}^{j_{0}} \sum_{i\in\overline{S}_{t}^{j}\backslash\overline{S}_{t}^{j-1}} {x_i(t) \over n_i(t)} \le \frac{\Delta_{x\left(t\right)}^2}{2\,f\!\left(t\right)\, g(m) } \sum_{j=1}^{j_{0}} { |\overline{S}_{t}^{j}\backslash\overline{S}_{t}^{j-1}| \over \alpha_j}
        \end{align*}
        The inner sum can be decomposed as follows, by definition of $S_{t}^{j}$ and $\overline{S}_{t}^{j}$:
        \begin{align*}
            \sum_{j=1}^{j_{0}}\frac{\left|\overline{S}_{t}^{j}\backslash\overline{S}_{t}^{j-1}\right|}{\alpha_{j}} & =\sum_{j=1}^{j_{0}}\frac{\left|S_{t}^{j}\backslash S_{t}^{j-1}\right|}{\alpha_{j}},\qquad\text{dropping the complements}\\
            & =\sum_{j=1}^{j_{0}}\frac{\left|S_{t}^{j}\right|-\left|S_{t}^{j-1}\right|}{\alpha_{j}}\\
            & =\frac{\left|S_{t}^{j_{0}}\right|}{\alpha_{0}}+\sum_{j=1}^{j_{0}}\left[\left|S_{t}^{j}\right|\left(\frac{1}{\alpha_{j-1}}-\frac{1}{\alpha_{j}}\right)\right],\qquad\text{by factoring the last term }j_{0}\\
            & <\frac{m\,\beta_{j_{0}}}{\alpha_{0}}+\sum_{j=1}^{j_{0}}\left[m\,\beta_{j}\left(\frac{1}{\alpha_{j-1}}-\frac{1}{\alpha_{j}}\right)\right],\qquad\text{as }\overline{A}_{t}\text{ holds}.
        \end{align*}
        Finally, replacing $g$ and $l$ by their definition
        \begin{align*}
            \sum_{i=1}^d {x_i(t) \over n_i(t)}  < \frac{m\,\Delta_{x\left(t\right)}^2}{2\,f\!\left(t\right)\,g(m)}\,\left(\frac{\beta_{j_{0}}}{\alpha_{j_{0}}}+\sum_{j=1}^{j_{0}}\frac{\beta_{j-1}-\beta_{j}}{\alpha_{j-1}}\right) = \frac{\Delta_{x\left(t\right)}^2 \overline{\varepsilon}_T^2}{8\,f\!\left(t\right)}
         \end{align*}
        We prove that event $\Delta_{x\left(t\right)} \le {4 \over \overline{\varepsilon}_T}  E_{t}(x(t))$ implies $A_{t}$. Indeed, if $\Delta_{x\left(t\right)} \le {4 \over \overline{\varepsilon}_T}  E_{t}(x(t))$ and $\overline{A}_{t}$, then
        \begin{align*}
            \Delta_{x\left(t\right)}^2
            \le {16 \over \overline{\varepsilon}_T^2}  E_{t}\!\left(x\!\left(t\right)\right)^2
            = {16 \over \overline{\varepsilon}_T^2} \sigma^2(t)^\top x(t)
            = {8 f(t) \over \overline{\varepsilon}_T^2} \sum_{i=1}^d {x_i \over n_i(t)}
            < \Delta_{x\left(t\right)}^2
        \end{align*}
        which is a contradiction. Therefore, $\Delta_{x\left(t\right)} \le {4 \over \overline{\varepsilon}_T}  E_{t}(x(t))$ implies $A_{t}$.

        We now bound the regret due to the event $A_t$. We further decompose $A_{t}^{j}$ to include the fact that a specific item $i$ is included among the (at least) $m\,\beta_{j}$ items that have not yet been selected enough:
        \begin{align*}
            A_{t}^{j,i}=A_{t}^{j}\cap\left\{ x_i(t) = 1 \, , \,n_i(t) \leq\frac{\alpha_{j}\,2\,f\!\left(T\right)\,\, g(m)}{\Delta_{x\left(t\right)}^{2}}\right\} .
        \end{align*}
        Of course, the union over all $i$ yields back $A_{t}^{j}$:
        \begin{align*}
            \bigcup_{i=1}^{d}A_{t}^{j,i}=A_{t}^{j}.
        \end{align*}
        Since $A_{t}^{j}$ implies that at least $m\,\beta_{j}$ items have not yet been selected enough,
        \begin{align*}
            \indic\left\{ A_{t}^{j} \right\} \leq\frac{1}{m\,\beta_{j}}\,\sum_{i=1}^{d}\indic \left\{ A_{t}^{j,i}\right\}.
        \end{align*}
        The contribution to the regret is bounded by the items that are not selected frequently enough to ensure a good reward estimate:

        \begin{align*}
            \sum_{t=1}^{T}\Delta_{x(t)}  \,\indic\left\{ F_t \right\} \leq\sum_{t=1}^{T}\Delta_{x\left(t\right)}\,\indic\left\{ A_{t}\right\} \leq \sum_{t=1}^{T}\sum_{j=1}^{+\infty}\Delta_{x\left(t\right)}\,\indic\left\{ A_{t}^{j}\right\} \leq \sum_{t=1}^{T}\sum_{j=1}^{+\infty}\sum_{i=1}^{d}\frac{\Delta_{x\left(t\right)}}{m\,\beta_{j}}\,\indic\left\{ A_{t}^{j,i}\right\} .
        \end{align*}
        For any $i$, define the possible values of the gaps $\Delta_x$ where $x_i = 1$, namely:
        \begin{align*}
            \{\Delta_{x} : x \in \cX , x_i = 1 \} = \{\Delta_{i,1},...,\Delta_{i,K_i}\}
        \end{align*}
        where $K_i$ is the number of possible values for the gap and we assume that the gaps are sorted in decreasing order:
        \begin{align*}
            \Delta_{i,1} > ... > \Delta_{i,K_i}
        \end{align*}
        with the convention that $\Delta_{i,0} = \infty$. We can then decompose the previous sum according to the value of the gap:
        \begin{align*}
            \sum_{t=1}^{T}\indic\!\left\{ F_{t} \right\} \Delta_{x\left(t\right)}
                & \le \sum_{t=1}^{T}\sum_{j=1}^{+\infty}\sum_{i=1}^{d}\frac{\Delta_{x\left(t\right)}}{m\,\beta_{j}}\,\indic\left\{ A_{t}^{j,i}\right\} \\
                & \leq \sum_{t=1}^{T}\sum_{j=1}^{+\infty}\sum_{i=1}^{d}\sum_{k=1}^{K_{i}}\frac{\Delta_{i,k}}{m\,\beta_{j}}\,\indic\left\{ A_{t}^{j,i},\Delta_{x\left(t\right)}=\Delta_{i,k}\right\} \\
                & \leq \sum_{t=1}^{T}\sum_{j=1}^{+\infty}\sum_{i=1}^{d}\sum_{k=1}^{K_{i}}\frac{\Delta_{i,k}}{m\,\beta_{j}}\indic\Bigg\{ A_{t}^{j}, x_i(t) = 1, n_{i}(t) \leq\frac{\alpha_{j}  f\!\left(T\right)\, g(m)}{2 \Delta_{i,k}^{2}}, \Delta_{x\left(t\right)}=\Delta_{i,k}\Bigg\},
        \end{align*}
        by definition of $A_{t}^{j,i}$. To simplify notation, let
        \begin{align*}
        \tau_{j}={1 \over 2} \alpha_{j} f(T) g(m).
        \end{align*}
        Thus, the previous bound can be written as:
        \begin{align*}
            \sum_{t=1}^{T}\indic\!\left\{ F_{t} \right\} \Delta_{x\left(t\right)}  \leq \sum_{t=1}^{T}\sum_{j=1}^{+\infty}\sum_{i=1}^{d}\sum_{k=1}^{K_{i}}\frac{\Delta_{i,k}}{m\,\beta_{j}}\indic\left\{ i:x_i(t) = 1 ,\quad n_{i}\!\left(t\right)\leq\frac{\tau_{j}}{\Delta_{i,k}^{2}},\quad\Delta_{x\left(t\right)}=\Delta_{i,k}\right\} .
        \end{align*}
        To simplify the developments, focus on the two sums, the one on the rounds $t$ and the one on the gaps $k$:
        \begin{align*}
            \sum_{t=1}^{T}\sum_{k=1}^{K_{i}}\frac{\Delta_{i,k}}{m\,\beta_{j}}\indic\left\{ x_i(t) = 1,\quad n_{i}\!\left(t\right)\leq\frac{\tau_{j}}{\Delta_{i,k}^{2}},\quad\Delta_{x\left(t\right)}=\Delta_{i,k}\right\}
        \end{align*}
        As the values of $\Delta_{i,k}$ are ordered, we can decompose as follows:
        \begin{align*}
            \sum_{t=1}^{T} &\sum_{k=1}^{K_{i}}\frac{\Delta_{i,k}}{m\,\beta_{j}}\indic\left\{ x_i(t)=1,\quad n_i(t)\leq\frac{\tau_{j}}{\Delta_{i,k}^{2}},\quad\Delta_{x\left(t\right)}=\Delta_{i,k}\right\}  \\
            &=\sum_{t=1}^{T}\sum_{k=1}^{K_{i}}\sum_{p=1}^{n}\frac{\Delta_{i,k}}{m\,\beta_{j}}\indic\left\{ x_i(t)=1,\quad n_i(t)\in\left(\frac{\tau_{j}}{\Delta_{i,p-1}^{2}},\frac{\tau_{j}}{\Delta_{i,p}^{2}}\right],\quad\Delta_{x\left(t\right)}=\Delta_{i,k}\right\}
        \end{align*}
        The factor $\Delta_{i,k}$ can be moved within the summation over $j$ and be rewritten as $\Delta_{i,k}$, as it will be counted only once, when the step function is nonzero (when $j=k$):
        \begin{align*}
            \sum_{t=1}^{T} & \sum_{k=1}^{K_{i}}\frac{\Delta_{i,k}}{m\,\beta_{j}}\indic\left\{ x_i(t)=1,\quad n_i(t)\leq\frac{\tau_{j}}{\Delta_{i,k}^{2}},\quad\Delta_{x\left(t\right)}=\Delta_{i,k}\right\}\\
            &\leq\sum_{t=1}^{T}\sum_{k=1}^{K_{i}}\sum_{p=1}^{n}\frac{\Delta_{i,p}}{m\,\beta_{j}}\indic\left\{ x_i(t)=1,\quad n_i(t)\in\left(\frac{\tau_{j}}{\Delta_{i,p-1}^{2}},\frac{\tau_{j}}{\Delta_{i,p}^{2}}\right],\quad\Delta_{x\left(t\right)}=\Delta_{i,k}\right\}
        \end{align*}
        If the sum over $p$ goes to $K_{i}$, many new terms can be added, though:
        \begin{align*}
            \sum_{t=1}^{T} &\sum_{k=1}^{K_{i}}\frac{\Delta_{i,k}}{m\,\beta_{j}}\indic\left\{ x_i(t)=1,\quad n_i(t)\leq\frac{\tau_{j}}{\Delta_{i,k}^{2}},\quad\Delta_{x\left(t\right)}=\Delta_{i,k}\right\} \\
            &\leq\sum_{t=1}^{T}\sum_{k=1}^{K_{i}}\sum_{p=1}^{K_{i}}\frac{\Delta_{i,p}}{m\,\beta_{j}}\indic\left\{ x_i(t)=1,\quad n_i(t)\in\left(\frac{\tau_{j}}{\Delta_{i,p-1}^{2}},\frac{\tau_{j}}{\Delta_{i,p}^{2}}\right],\quad\Delta_{x\left(t\right)}=\Delta_{i,k}\right\}
        \end{align*}
        Again, if the solution $x\!\left(t\right)$ is not taken to be exactly $k$, but any suboptimal solution, many new terms now count in the summation. With this change, the sum over $k$ becomes irrelevant, as all gaps that may contribute to the regret are still counted in.
        \begin{align*}
            \sum_{t=1}^{T}& \sum_{k=1}^{K_{i}}\frac{\Delta_{i,k}}{m\,\beta_{j}}\indic\left\{ x_i(t)=1,\quad n_i(t)\leq\frac{\tau_{j}}{\Delta_{i,k}^{2}},\quad\Delta_{x\left(t\right)}=\Delta_{i,k}\right\} \\
            &\leq\sum_{t=1}^{T}\sum_{p=1}^{K_{i}}\frac{\Delta_{i,p}}{m\,\beta_{j}}\indic\left\{ x_i(t)=1,\quad n_i(t)\in\left(\frac{\tau_{j}}{\Delta_{i,p-1}^{2}},\frac{\tau_{j}}{\Delta_{i,p}^{2}}\right],\quad\Delta_{x\left(t\right)}>0\right\} \\
            &\leq \frac{\tau_j}{m\,\beta_{j}} \left({1 \over \Delta_{i,1}} + \sum_{p=2}^{K_{i}}  \Delta_{i,p} \left( { 1\over \Delta_{i,p}^2} -  {1 \over \Delta_{i,p-1}^2} \right)   \right) \le \frac{2 \tau_j}{m\,\beta_{j} \Delta_{\min}}
        \end{align*}
        where we used the following algebra, since $ \Delta_{i,1} > ...  > \Delta_{i,K_i}$:
        \begin{align*}
            {1 \over \Delta_{i,1}} + \sum_{p=2}^{K_{i}}  \Delta_{i,p} \left( { 1\over \Delta_{i,p}^2} -  {1 \over \Delta_{i,p-1}^2} \right) = {1 \over \Delta_{i,K_i}} + \sum_{p=1}^{K_{i}-1} {\Delta_{i,p} - \Delta_{i,p+1} \over \Delta_{i,p}^2}  \le {1 \over \Delta_{i,K_i}} + \sum_{p=1}^{K_{i}-1} {\Delta_{i,p} - \Delta_{i,p+1} \over \Delta_{i,p+1} \Delta_{i,p}} \\
            =  {1 \over \Delta_{i,K_i}} + \sum_{p=1}^{K_{i}-1} {1 \over \Delta_{i,p+1}} - {1 \over \Delta_{i,p}} = {2 \over \Delta_{i,K_i}} - {1 \over \Delta_{i,1}} \le {2 \over \Delta_{\min}}
        \end{align*}
        Injecting this result into the regret term bound,
        \begin{align*}
            \sum_{t=1}^{T}\indic\!\left\{ F_{t} \right\} \Delta_{x\left(t\right)}  & \leq  \sum_{t=1}^{T}\sum_{j=1}^{+\infty}\sum_{i=1}^{d}\sum_{k=1}^{K_{i}}\frac{\Delta_{i,k}}{m\,\beta_{j}}\indic\left\{ x_i(t)=1,\quad n_i(t)\leq\frac{\tau_{j}}{\Delta_{i,k}^{2}},\quad\Delta_{x\left(t\right)}=\Delta_{i,k}\right\} \\
             & \leq \sum_{j=1}^{+\infty}\sum_{i=1}^{d} \frac{2 \tau_j}{m\,\beta_{j} \Delta_{\min}} \\
             & =  \frac{f\!\left(T\right) d  g(m)}{m \Delta_{\min}}  \left[ \sum_{j=1}^{j_{0}}\frac{\alpha_{j}}{\beta_{j}} \right],\qquad\text{by definition of }\tau_{j}\\
             & =\frac{4 l d \,f\left(T\right)}{\overline{\varepsilon}_T^2 \Delta_{\min}}\left[ \sum_{j=1}^{j_{0}}\frac{\alpha_{j}}{\beta_{j}} \right] ,\qquad\text{by definition of }g.
        \end{align*}
        Now, set $\alpha_{i}=\beta_{i}=\beta^{i},\qquad\beta\in\left(0,1\right)$, which satisfies the previous assumptions. Since $j_{0}$ is the first integer $j$ such that $\beta_{j}\leq m^{-1}$, we have $j_{0}=\left\lceil \frac{\ln m}{\ln\beta^{-1}}\right\rceil$. Also,
        \begin{align*}
            l \sum_{j=1}^{j_{0}}\frac{\alpha_{j}}{\beta_{j}} & =l \,j_{0},\qquad\text{as }\alpha_{i}=\beta_{i}\\
             & =j_{0}\left(\frac{\beta_{j_{0}}}{\alpha_{j_{0}}}+\sum_{j=1}^{j_{0}}\frac{\beta_{j-1}-\beta_{j}}{\alpha_{j-1}}\right),\qquad\text{by definition of }l\\
             & =j_{0}\left(1+\sum_{j=1}^{j_{0}}\frac{\beta^{j-1}-\beta^{j}}{\beta^{j}}\right)\\
             & =j_{0}\left(1+\sum_{j=1}^{j_{0}}\frac{1-\beta}{\beta}\right)\\
             & =j_{0}\left(1+\frac{j_{0}}{\beta}-j_{0}\right)\\
             & \leq j_{0}(1 + \frac{j_{0}}{\beta}).
        \end{align*}
        Taking $\beta=1/5$,
        \begin{align*}
            j_{0}=\left\lceil \frac{\ln m}{\ln\beta^{-1}}\right\rceil \leq\left\lceil \frac{\ln m}{1.61}\right\rceil .
        \end{align*}
        Injecting these into the regret term, we get:
        \begin{align*}
            \sum_{t=1}^{T}\Delta_{x(t)} \indic\left\{ F_t \right\}  & \leq \frac{4 l d \,f\left(T\right)}{\overline{\varepsilon}_T^2 \Delta_{\min}}\left[ \sum_{j=1}^{j_{0}}\frac{\alpha_{j}}{\beta_{j}} \right] \leq \frac{4  d \,f\left(T\right)}{\overline{\varepsilon}_T^2 \Delta_{\min}}  \left(\left\lceil \frac{\ln m}{1.61}\right\rceil +5\left\lceil \frac{\ln m}{1.61}\right\rceil ^{2}\right) \leq \frac{24  d \,f\left(T\right)}{\overline{\varepsilon}_T^2 \Delta_{\min}} \left\lceil \frac{\ln m}{1.61}\right\rceil ^{2}.
        \end{align*}

    \paragraph{Complete regret bound.}

        Gathering the results about the three terms of the regret decomposition, the regret can be bounded by:
        \begin{align*}
            R\!\left(T\right)\leq C_4(m) + \frac{2\,d\,m^{3}}{\Delta_{\min}^{2}}+ \frac{24  d \,f\left(T\right)}{\overline{\varepsilon}_T^2 \Delta_{\min}} \left\lceil \frac{\ln m}{1.61}\right\rceil ^{2} + 4 \sum_{t=1}^{T} \delta_t \indic\!\left\{ \Delta_{\min} \le 4 \delta_t \right\}.
        \end{align*}

\section{Proof of Theorem 5.1}

    The first step is to upper bound the value of problem $(P_2)$ with inputs $a(t)$ and $b(t)$ as a function of $x(t)$.

    Define the set:
    \begin{align*}
        S(t) = \{0,...,m \xi(t)\}.
    \end{align*}
    By definition, $a(t) \in     \{0,...,\xi(t)\}$, and $m = \max_{x \in \cX} \indic^\top x$, so:
    \begin{align*}
        a(t)^\top x \in  S(t) , \forall x \in \cX.
    \end{align*}
    Therefore:
    \begin{align*}
        \max_{x \in \cX} & \Big \{ a(t)^\top x  + \sqrt{b(t)^\top x} \Big\} \\
        &= \max_{s \in S} \max_{x \in X, a(t)^\top x = s} \left \{ s + \sqrt{b(t)^\top x} \right\} \\
                &\le \max_{s \in S} \max_{x \in X, a(t)^\top x \ge s} \left \{ s + \sqrt{b(t)^\top x} \right\} \\
                &\le \max_{s \in S} \left\{ s  + {1 \over \varepsilon_t} \sqrt{b(t)^\top \overline{x}^s(t)} \right\} \\
                &= s^\star(t) +  {1 \over \varepsilon_t} \sqrt{b(t)^\top \overline{x}^{s^\star(t)}(t)} \\
                &\le a(t)^\top \overline{x}^{s^\star(t)}(t) +  {1 \over \varepsilon_t} \sqrt{b(t)^\top \overline{x}^{s^\star(t)}(t)} \\
                &= a(t)^\top x(t)  +  {1 \over \varepsilon_t} \sqrt{b(t)^\top x(t)} \\
    \end{align*}
    where we used the fact that, by definition, $\overline{x}^s(t)$ is an $\varepsilon_t$-approximate solution to $(P_3)$, that
    \begin{align*}
        s^\star(t) \in \arg \max_{s \in S(t)} \left\{ s  + {1 \over \varepsilon_t} \sqrt{b(t)^\top \overline{x}^s(t)} \right\}
    \end{align*}
    and that $a(t)^\top \overline{x}^{s}(t) \ge s$ for all $s \in S(t)$.

    The second step is to relate the value of problem $(P_2)$ with inputs $a(t)$ and $b(t)$ to the value of problem $(P_2)$ with inputs $\hat\theta(t)$ and $\sigma^2(t)$. We recall that, by definition, we have
    \begin{align*}
        a_i(t) &= \lceil{ \xi(t) \hat\theta_i(n) } \rceil, i=1,...,d\\
        b_i(t) &= \xi^2(t) \sigma^2(t), i=1,...,d
    \end{align*}
    where $\xi(t) = \lceil {m \over \delta_t} \rceil$. Therefore, $a(t) \in \{1,...,\xi(t)\}^{d}$ and:
    \begin{align*}
        \hat\theta(t) \le {1 \over \xi(t)} a(t) \le {1 \over \xi(t)} \indic + \hat\theta(t)
    \end{align*}
    As a consequence, for any $x \in \cX$:
    \begin{align*}
         \hat\theta(t)^\top x \le {1 \over \xi(t)} a(t)^\top x \le {1 \over \xi(t)} \indic^\top x  + \hat\theta(t)^\top x \le \delta_t + \hat\theta(t)^\top x
    \end{align*}
    We have proven in the first step that:
    \begin{align*}
        \max_{x \in \cX} & \Big \{ a(t)^\top x  + \sqrt{b(t)^\top x} \Big\} \\
        &\le a(t)^\top x(t)  +  {1 \over \varepsilon_t} \sqrt{b(t)^\top x(t)}
    \end{align*}
    Then:
    \begin{align*}
        \max_{x \in \cX} & \{ \hat\theta(t)^\top x  + \sqrt{\sigma^2(t)^\top x } \} \\
        &= {1 \over \xi(t)} \max_{x \in \cX}\{ \xi(t)\hat\theta(t)^\top x   + \sqrt{\xi(t)^2 \sigma^2(t) x } \} \\
        &\le {1 \over \xi(t)} \max_{x \in \cX}\{ a(t)^\top x  + \sqrt{b(t)^\top x } \} \\
        &\le {1 \over \xi(t)} a(t)^\top x(t)  + {1 \over \xi(t) \varepsilon_t} \sqrt{b(t)^\top x(t)  } \\
        &\le \delta_t +  \hat\theta(t)^\top  x(t)+ {1 \over \varepsilon_t} \sqrt{\sigma^2(t)^\top x(t) }.
    \end{align*}
    which is the announced result.

\section{Pseudo-code to solve problem $(P_2)$}

    In this section, we provide the pseudo-code for the algorithms described in Section~5, for each family of combinatorial sets. We make use of several subroutines: ${\tt Hungarian}$ is the Hungarian algorithm computing a maximum weighted matching in a bipartite graph, ${\tt Greedy}$ is the greedy algorithm finding a maximum weighted spanning tree of a graph, ${\tt Meggido}$ is Meggido's search algorithm to minimize a piecewise-linear function, ${\tt Bellman}$ is the Bellman-Ford algorithm for the shortest path in a directed acyclic graph.

    \begin{algorithm}[p]
        \SetKwInOut{Input}{input}

        \caption{Algorithm for the budgeted $m$-set problem.}

        \Input{Number $m\in\{1,\dots,d\}$, rewards $b\in\mathds{R}^{|E|}$, weights $a\in\mathds{N}^{|E|}$}
        \KwResult{Solution to $P_3(s)$ for all $s\in\{0,1,\dots,d\max_{i}a_i\}$}
        $L\gets\text{empty array of size }(d\max_{i}a_i+1,m,d)$\\
        $S\gets\text{empty array of size }(d\max_{i}a_i+1,m,d)$\\
        $S^\star\gets\text{empty array of size }(d\max_{i}a_i+1)$
        \For{$s=0,1,\dots,d\max_{i}a_i$}{
            \For{$\ell=0,1,\dots,m$}{
                \For{$i=d,d-1,\dots,0$}{
                    \eIf{$i = d$}{
                        \eIf{$s = 0$}{
                            $L[s,\ell,i]\gets0$\\
                            $S[s,\ell,i]\gets\emptyset$
                        }{
                            $L[s,\ell,i]\gets-\infty$\\
                            $S[s,\ell,i]\gets\nexists$
                        }
                    }{
                        \eIf{$\ell=0$}{
                            $L[s,\ell,i]\gets L[s,\ell,i+1]$\\
                            $S[s,\ell,i]\gets S[s,\ell,i+1]$
                        }{
                            $L[s,\ell,i]\gets\max\{b_i+L[\max(s-a_i,0),\ell,i+1],L[s,\ell,i+1]\}$
                            \eIf{$L[s,\ell,i]=,L[s,\ell,i+1]$}{
                                $S[s,\ell,i]\gets S[s,\ell,i+1]$
                            }{
                                $S[s,\ell,i]\gets S[s,\ell,i+1]\cup\{i\}$
                            }
                        }
                    }
                }
            }
            $S^\star[s]\gets S[s,m,0]$
        }
        \Return$S^\star$
    \end{algorithm}
    \begin{algorithm}[p]
        \SetKwInOut{Input}{input}

        \caption{Algorithm for the budgeted source-destination path problem.}

        \Input{A directed graph $G=(V,E)$, source node $u\in V$, destination node $v\in V$, length of the longest path $m$, rewards $b\in\mathds{R}^{|E|}$, weights $a\in\mathds{N}^{|E|}$}
        \KwResult{Solution to $P_3(s)$ for all $s\in\{0,1,\dots,m\max_{i}a_i\}$}
        $L\gets\text{empty array of size }(m\max_{i}a_i+1,|V|)$\\
        $S\gets\text{empty array of size }(m\max_{i}a_i+1,|V|)$\\
        $S^\star\gets\text{empty array of size }(d\max_{i}a_i+1)$\\
        \For{$s=0,1,\dots,d\max_{i}a_i$}{
            \For{$w \in V$}{
                \eIf{$s = 0$}{
                    $L[s,w], S[s,w]\gets {\tt Bellman}(G,b,u,w)$
                }{
                    $x^\star\gets\arg\max_{x:(w,x)\in E}\{b_{(w,x)}+L[x,\max(s-a_{(w,x)},0)]\}$\\
                    $L[s,w]\gets b_{(w,x^\star)}+L[x^\star,\max(s-a_{(w,x^\star)},0)]$\\
                    $S[s,w]\gets (w,x)\cup S[x^\star,\max(s-a_{(w,x^\star)},0)]$
                }
            }
            $S^\star[s]\gets S[s,v]$
        }
        \Return$S^\star$
    \end{algorithm}
    \begin{algorithm}[p]
        \SetKwInOut{Input}{input}

        \caption{Approximation algorithm for the budgeted maximum spanning tree problem.}

        \Input{An undirected graph $G=(V,E)$, rewards $b\in\mathds{R}^{|E|}$, weights $a\in\mathds{N}^{|E|}$, a minimum budget $s\in\mathds{N}$}
        \KwResult{Solution to $P_3(s)$}
        $x\gets nothing$\\
        \For{all unordered pairs $(e_1,e_2)$ of distinct edges of $E$}{
            $E'\gets\{e\in E\ |\ b_e\leq\min\{b_{e_1},b_{e_2}\}\}$\\
            $G'\gets(V,E')$\\
            $x^\star,\lambda^\star\gets {\tt Meggido}({\tt Greedy},G',a+\lambda b)$\\
            $\varepsilon\gets\text{arbitrary small value}$\\
            $x^+\gets {\tt Greedy}(G',a+(\lambda^\star+\varepsilon)b)$\\
            $x^-\gets {\tt Greedy}(G',a+(\lambda^\star-\varepsilon)b)$\\
            \While{$|x^+\oplus x^-|>1$}{
                Find $e$, $e'$ such that $x^+_e=x^-_{e'}=1$ and $x^+_{e'}=x^-_e=0$\\
                $\tilde x\gets x^+-e_e+e_{e'}$\\
                \eIf{$a^\top\tilde x\ge s$}{
                    $x^+\gets \tilde x$
                }{
                    $x^-\gets \tilde x$
                }
            }
            \If{$x$ is nothing \normalfont{\textbf{or}} $b^\top\tilde x>b^\top x$}{
                $x\gets x^+$
            }
        }
        \Return$x$
    \end{algorithm}
    \begin{algorithm}[p]
        \SetKwInOut{Input}{input}

        \caption{Approximation algorithm for the budgeted maximum bipartite matching problem.}

        \Input{A bipartite graph $G=(V,E)$, rewards $b\in\mathds{R}^{|E|}$, weights $a\in\mathds{N}^{|E|}$, a minimum budget $s\in\mathds{N}$}
        \KwResult{Solution to $P_3(s)$}
        $x\gets nothing$\\
        \For{all unordered 4-tuples $(e_1,e_2,e_3,e_4)$ of distinct edges of $E$}{
            $E'\gets\{e\in E\ |\ b_e\leq\min\{b_{e_1},b_{e_2},b_{e_3},b_{e_4}\}\}$\\
            $G'\gets(V,E')$\\
            $x^\star,\lambda^\star\gets {\tt Meggido}({\tt Hungarian},G',a+\lambda b)$\\
            $\varepsilon\gets\text{arbitrary small value}$\\
            $x^+\gets {\tt Hungarian}(G',a+(\lambda^\star+\varepsilon)b)$\\
            $x^-\gets {\tt Hungarian}(G',a+(\lambda^\star-\varepsilon)b)$\\
            \While{$|x^+\oplus x^-|>2$}{
                $x'\gets x^+\oplus x^-$\\
                $x''\gets\text{one path or one cycle from }x'$\\
                $\tilde x\gets x^-\oplus x''$\\
                \eIf{$a^\top\tilde x\ge s$}{
                    $x^+\gets \tilde x$
                }{
                    $x^-\gets \tilde x$
                }
            }
            \If{$x$ is nothing \normalfont{\textbf{or}} $b^\top\tilde x>b^\top x$}{
                $x\gets x^+$
            }
        }
        \Return$x$
    \end{algorithm}

\newpage
\bibliography{main}

\begin{thebibliography}{10}

\bibitem{abbasi11}
Yasin Abbasi-Yadkori, D\'{a}vid P\'{a}l, and Csaba Szepesv\'{a}ri.
\newblock Improved algorithms for linear stochastic bandits.
\newblock In {\em Proc. of NIPS}. 2011.

\bibitem{anantharam1987asymptotically_iid}
Venkatachalam Anantharam, Pravin Varaiya, and Jean Walrand.
\newblock Asymptotically efficient allocation rules for the multiarmed bandit
  problem with multiple plays-part i: iid rewards.
\newblock {\em Automatic Control, IEEE Transactions on}, 32(11):968--976, 1987.

\bibitem{atamturk2017}
Alper Atamt{\"{u}}rk and Andr{\'{e}}s G{\'{o}}mez.
\newblock Maximizing a class of utility functions over the vertices of a
  polytope.
\newblock {\em Operations Research}, 65(2):433--445, 2017.

\bibitem{auer2003}
Peter Auer, Nicol\`{o} Cesa-Bianchi, Yoav Freund, and Robert~E. Schapire.
\newblock The nonstochastic multiarmed bandit problem.
\newblock {\em SIAM J. Comput.}, 32(1):48--77, 2003.

\bibitem{berger2011}
Andre Berger, Vincenzo Bonifaci, Fabrizio Gandoni, and Guido Shaefer.
\newblock Budgeted matching and budgeted matroid intersection via the gasoline
  puzzle.
\newblock {\em Mathematical Programming}, 2011.

\bibitem{julia}
Jeff. Bezanson, Alan. Edelman, Stefan. Karpinski, and Viral~B. Shah.
\newblock Julia: A fresh approach to numerical computing.
\newblock {\em SIAM Review}, 59(1):65--98, 2017.

\bibitem{cappe2012}
Olivier Capp\'e, Aurelien Garivier, Odalric Maillard, Remi Munos, and Gilles
  Stoltz.
\newblock Kullback-leibler upper confidence bounds for optimal sequential
  allocation.
\newblock {\em Annals of Statistics}, 41(3):516--541, June 2013.

\bibitem{cesa2012}
Nicol{\`o} Cesa-Bianchi and G{\'a}bor Lugosi.
\newblock Combinatorial bandits.
\newblock {\em Journal of Computer and System Sciences}, 78(5):1404--1422,
  2012.

\bibitem{chen2013}
Wei Chen, Yajun Wang, and Yang Yuan.
\newblock Combinatorial multi-armed bandit: General framework, results and
  applications.
\newblock In {\em Proc. of ICML}, 2013.

\bibitem{chu11}
Wei Chu, Lihong Li, Lev Reyzin, and Robert Schapire.
\newblock Contextual bandits with linear payoff functions.
\newblock In {\em Proc. of AISTATS}, 2011.

\bibitem{coey2018}
Chris Coey, Miles Lubin, and Juan~Pablo Vielma.
\newblock Outer approximation with conic certificates for mixed-integer convex
  problems.
\newblock {\em Math. Program. Comput.}, 12(2):249--293, 2020.

\bibitem{combes2015}
Richard Combes, M.~Sadegh Talebi, Alexandre Proutiere, and Marc Lelarge.
\newblock Combinatorial bandits revisited.
\newblock In {\em Proc. of NIPS}, 2015.

\bibitem{dani08}
Varsha Dani, Thomas~P. Hayes, and Sham~M. Kakade.
\newblock Stochastic linear optimization under bandit feedback.
\newblock In {\em Proc. of COLT}, pages 355--366, 2008.

\bibitem{degenne2016}
Remy Degenne and Vianney Perchet.
\newblock Combinatorial semi-bandit with known covariance.
\newblock In {\em Proc. of NIPS}, 2016.

\bibitem{kveton2014}
Branislav Kveton, Zheng Wen, Azin Ashkan, Hoda Eydgahi, and Brian Eriksson.
\newblock Matroid bandits: Fast combinatorial optimization with learning.
\newblock In {\em Proc. of UAI}, 2014.

\bibitem{kveton2014tight}
Branislav Kveton, Zheng Wen, Azin Ashkan, and Csaba Szepesvari.
\newblock Tight regret bounds for stochastic combinatorial semi-bandits.
\newblock In {\em Proc. of AISTATS}, 2015.

\bibitem{lai1985}
Tze~Leung Lai and Herbert Robbins.
\newblock Asymptotically efficient adaptive allocation rules.
\newblock {\em Advances in Applied Mathematics}, 6(1):4--2, 1985.

\bibitem{lobo1998}
Miguel~Sousa Lobo, Lieven Vandenberghe, Stephen Boyd, and Hervé Lebret.
\newblock Applications of second-order cone programming.
\newblock {\em Linear Algebra and its Applications}, 284(1):193 -- 228, 1998.

\bibitem{meggido1981}
Nimrod Megiddo.
\newblock Applying parallel computation algorithms in the design of serial
  algorithms.
\newblock In {\em Proc. of FOCS}, 1981.

\bibitem{oxley2006}
James~G. Oxley.
\newblock {\em Matroid Theory (Oxford Graduate Texts in Mathematics)}.
\newblock Oxford University Press, Inc., USA, 2006.

\bibitem{perrault2019}
Pierre Perrault, Vianney Perchet, and Michal Valko.
\newblock Exploiting structure of uncertainty for efficient matroid
  semi-bandits.
\newblock In {\em Proc. of ICML}, 2019.

\bibitem{ravi1996}
R.~Ravi and Michel~X. Goemans.
\newblock The constrained minimum spanning tree problem.
\newblock {\em SWAT}, 1996.

\bibitem{rejwan20}
Idan Rejwan and Yishay Mansour.
\newblock Top-$k$ combinatorial bandits with full-bandit feedback.
\newblock Proc. of ALT, 2020.

\bibitem{robbins1952}
Herbert Robbins.
\newblock Some aspects of the sequential design of experiments.
\newblock {\em Bulletin of the American Mathematical Society}, 58(5):527--535,
  1952.

\bibitem{talebi2016}
M.~Sadegh Talebi and Alexandre Proutiere.
\newblock An optimal algorithm for stochastic matroid bandit optimization.
\newblock In {\em Proc. of ICAAMS}, 2016.

\bibitem{wang2018}
Siwei Wang and Wei Chen.
\newblock Thompson sampling for combinatorial semi-bandits.
\newblock In {\em Proc. of ICML}, 2018.

\bibitem{wen2015}
Zheng Wen, Branislav Kveton, and Azin Ashkan.
\newblock Efficient learning in large-scale combinatorial semi-bandits.
\newblock In {\em Proc. of ICML}, 2015.

\end{thebibliography}
\bibliographystyle{plain}
\end{document}